%% file: tpami.tex
\documentclass[10pt,journal,compsoc]{IEEEtran}

\usepackage{epsfig,graphicx,amsfonts,xcolor,multirow,amsmath,hyperref,booktabs,balance}
\usepackage[nocompress]{cite}
\usepackage{mycommand}
\usepackage{algorithm}
\usepackage{algorithmic}

\input{math_commands.tex}

\usepackage{url}
\usepackage{array}
\usepackage{microtype}
\usepackage{graphicx}
\usepackage{xcolor}
\usepackage{subfigure}
\usepackage{amsmath}
\usepackage{bm}
\usepackage{dsfont}
\usepackage{mathtools}
\usepackage{nccmath}
\usepackage{amssymb}
\usepackage{multirow}
\usepackage{booktabs} 
\usepackage{multicol}
\usepackage{varwidth}
\usepackage{pifont}
\usepackage{makecell}
\usepackage{wrapfig}
\usepackage[flushleft]{threeparttable}
\usepackage{varwidth}
\usepackage{pifont}
\usepackage{makecell}
\usepackage{enumitem}
\usepackage{adjustbox}
\usepackage{graphicx,calc}

\usepackage[textsize=scriptsize]{todonotes}

\DeclarePairedDelimiterX{\inp}[2]{\langle}{\rangle}{#1, #2}

\DeclareMathAlphabet\mathbfcal{OMS}{cmsy}{b}{n}

\definecolor{darkred}{RGB}{192, 0, 0}

\usepackage{environ}
\usepackage{amsthm}
\usepackage{graphicx} 
\usepackage{tikz}
\usepackage{wrapfig}
\usepackage{makecell}
\usepackage{diagbox}
\usepackage{pifont}
\usepackage{diagbox}

\NewEnviron{elaboration}{
\begin{tikzpicture}
\node[rectangle,minimum width=0.46\textwidth] (m) {\begin{minipage}{0.46\textwidth}\BODY\end{minipage}};
\draw[thick] (m.south west) rectangle (m.north east);
\end{tikzpicture}
}

\PassOptionsToPackage{hyphens}{url}\usepackage{hyperref}
\hypersetup{colorlinks=true}

\definecolor{Tianlong_color}{rgb}{0.858, 0.188, 0.478}

\def\mE{\mathbb{E}}
\def\mR{\mathbb{R}}
\def\bx{{\bf x}_i}
\def\bX{{\bf X}}
\def\cx{{\bf x}_{(k)}}
\def\tilx{\tilde{\bf x}_i}
\def\tilX{\tilde{\bf X}}
\def\std{\text{std}}
\def\bU{{\bf U}}
\newcommand{\be}{\begin{equation}}
\newcommand{\ee}{\end{equation}}

\begin{document}

\title{Bag of Tricks for Training Deeper Graph Neural Networks: A Comprehensive Benchmark Study}

\author{
  Tianlong~Chen$^\ddagger$,
  Kaixiong~Zhou$^\ddagger$,
  Keyu~Duan,
  Wenqing~Zheng,
  Peihao~Wang,
  Xia~Hu,
  and~Zhangyang~Wang

  \IEEEcompsocitemizethanks{
    \IEEEcompsocthanksitem Tianlong Chen, Wenqing Zheng, Peihao Wang and Zhangyang Wang are with the Department of Electrical and Computer Engineering, The University of Texas at Austin, TX, 78712. E-mail: \{tianlong.chen, w.zheng, peihaowang, atlaswang\}@utexas.edu
    \IEEEcompsocthanksitem Kaixiong Zhou, Keyu Duan and Xia Hu are with the Department of Computer Science, Rice University. E-mail: \{kz34, k.duan, xia.hu\}@rice.edu
    \IEEEcompsocthanksitem $\ddagger$ indicates the equal contribution. Correspondence to Zhangyang Wang (atlaswang@utexas.edu).
  }
}

\markboth{IEEE Transactions on Pattern Analysis and Machine Intelligence}
{Chen \MakeLowercase{\textit{et al.}}: Bag of Tricks for Training Deeper Graph Neural Networks: A Comprehensive Benchmark Study}

\IEEEtitleabstractindextext{
  \begin{abstract}
    Training deep graph neural networks (GNNs) is notoriously hard. Besides the standard plights in training deep architectures such as vanishing gradients and overfitting, it also uniquely suffers from over-smoothing, information squashing, and so on, which limits their potential power for encoding the high-order neighbor structure in large-scale graphs. Although numerous efforts are proposed to address these limitations, such as various forms of skip connections, graph normalization, and random dropping, it is difficult to disentangle the advantages brought by a deep GNN architecture from those ``tricks" necessary to train such an architecture. Moreover, the lack of a standardized benchmark with fair and consistent experimental settings poses an almost insurmountable obstacle to gauge the effectiveness of new mechanisms. In view of those, we present the first fair and reproducible benchmark dedicated to assessing the ``tricks" of training deep GNNs. We categorize existing approaches, investigate their hyperparameter sensitivity, and unify the basic configuration. Comprehensive evaluations are then conducted on tens of representative graph datasets including the recent large-scale Open Graph Benchmark, with diverse deep GNN backbones. We demonstrate that an organic combo of initial connection, identity mapping, group and batch normalization attains the new state-of-the-art results for deep GNNs on large datasets. Codes are available:  \url{https://github.com/VITA-Group/Deep_GCN_Benchmarking}.
  \end{abstract}

}

\maketitle
\IEEEpeerreviewmaketitle

\IEEEraisesectionheading{
  \section{Introduction}
  \label{sec:introduction}
}

\IEEEPARstart{G}{raph} neural networks (GNNs)~\cite{li2015gated} are powerful tools for modeling graph-structured data, and have been widely adopted in various real-world scenarios, including inferring individual relations in social and academic networks~\cite{tang2009relational,ying2018graph,zhou2019auto,gao2018large,you2020l2,you2020graph,you2022bringing,zheng2022cold,pmlr-v119-you20a}, improving predictions of recommendation systems~\cite{monti2017geometric,ying2018graph, zheng2021cold}, augmenting signal processing and communication~\cite{ortega2018graph,hu2021scalable,yang2016large}, modeling proteins for drug discovery~\cite{zitnik2017predicting,wale2008comparison}, segmenting large point clouds~\cite{wang2019dynamic,li2019deepgcns}, 
among others. While many classical GNNs have no more than just a few layers, recent advances attempt to investigate deeper GNN architectures~\cite{li2019deepgcns,li2021deepgcns, duan2018benchmarking}. There are massive examples on which deeper GNNs help, such as ``geometric" graphs representing structures of molecules, point clouds \cite{li2019deepgcns}, or meshes \cite{gong2020geometrically}. Deeper architectures are also found with superior performance on large-scale graphs like the latest Open Graph Benchmark (OGB)~\cite{hu2020open}.

However, training deep GNNs is notoriously challenging \cite{li2019deepgcns}.  Besides the standard plights in training deep architectures such as \textit{vanishing gradients} and \textit{overfitting}, the training of deep GNNs suffers several unique barriers that limit their potential power on the large-scale graphs. One of them is \textit{over-smoothing}, i.e., the node features tending to become indistinguishable as the result of performing several recursive neighborhood aggregation \cite{oono2020graph,li2018deeper,nt2019revisiting,wang2022anti}. This behavior was first observed in GCN models \cite{li2018deeper,nt2019revisiting}, which acts similarly to low-pass filters, and prevents deep GNNs from effectively modeling the higher-order dependencies from multi-hop neighbors. Another phenomenon is the \textit{bottleneck} phenomenon: while deep GNNs are expected to utilize longer-range interactions, the structure of graph often results in the exponential growth of receptive field, causing ``over-squashing” of information from the exponentially increasing neighbours into fixed-size vectors \cite{alon2020bottleneck}, and explaining why some deep GNNs do not improve in performance compared to their shallower peers. 

To address the aforementioned roadblocks, typical approaches could be categorized as \textit{architectural modifications}, and \textit{regularization \& normalization}: both we view as ``training tricks". The former includes various types of residual connections~\cite{li2019deepgcns,chen2020simple,klicpera2018predict,zhang2020revisiting,li2018deeper,klicpera2018predict,zhang2020revisiting,xu2018representation,liu2020towards}; and the latter include random edge dropping~\cite{rong2020dropedge,dropedge2}, pairwise distance normalization between node features (PairNorm) \cite{zhao2019pairnorm}, node-wise mean and variance normalization (NodeNorm) \cite{zhou2020understanding}, among many other normalization options ~\cite{ioffe2015batch,ba2016layer,wu2018group,yang2020revisiting,wang2022anti,zhou2020towards}.  While these techniques in general contribute to effective training of deep GNNs with tens of layers, their gains are not always significant nor consistent \cite{zhou2020understanding}. Furthermore, it is often non-straightforward to disentangle the gain by deepening the GNN architecture, from the ``tricks" necessary to train such a deeper architecture. In some extreme situations, contrary to the initial belief, newly proposed training techniques could improve shallower GNNs even to outperform deep GNNs \cite{zhou2020understanding}, making our pursuit of depth unpersuasive. Those observations have shed light on a \textbf{missing key knob} on studying deep GNNs: we lack a standardized benchmark that could offer fair and consistent comparison on the effectiveness of deep GNN training techniques. Without isolating the effects of deeper architectures from their training ``tricks", one might never reach convincing answers whether deep graph neural networks with \textit{ceteris paribus} should perform better. 
\vspace{-1em}
\subsection{Our Contributions}
Aiming to establish such a fair benchmark, our first step is to thoroughly investigate the design philosophy and implementation details on dozens of popular deep GNN training techniques, including various residual connections, graph normalization, and random dropping. The summarization could be found in Tables~\ref{tab:settings_cora},~\ref{tab:settings_cite},~\ref{tab:settings_pubmed}, and~\ref{tab:settings_arxiv}. Somehow unfortunately, we find that even sticking to the same dataset and GNN backbone, the hyperparameter configurations (e.g., hidden dimension, learning rate, weight decay, dropout rate, training epochs, early stopping patience) are highly inconsistently implemented, often varying case-to-case, which make it troubling to draw any fair conclusion.

To this end, in this paper we carefully examine those sensitive hyperparameters and unify them into one ``sweetpoint" hyperparameter set, to be adopted by all experiments. Such lays the foundation for a fair and reproducible benchmark of training deep GNN. Then, we comprehensively explore the diverse combinations of the available training techniques, over \textit{tens of} classical graph datasets with commonly used deep GNN backbones. Our comprehensive study turns out to be worthy. We conclude the baseline training configurations with the superior combo of training tricks, that lead us to attaining the new state-of-the-art results across multiple representative graph datasets including OGB. Specifically, we show that an organic combination of  initial connection, identity mapping, group and batch normalizations has preferably strong performance on large datasets.

Our experiments also reveals a number of ``surprises". For example, \ding{182} as we empirically show, while initial connection~\cite{chen2020simple} and jumping connection~\cite{xu2018representation} are both ``beneficial" training tricks when applied alone, combining them together deteriorates deep GNN performance. \ding{183} Although dense connection brings considerable improvement on large-scale graphs with deep GNNs, it sacrifices the training stability to a severe extent. \ding{184} As another example, the gain from NodeNorm~\cite{zhou2020understanding} becomes diminishing when applied to the large-scale datasets or deeper GNN backbones. \ding{185} Moreover, using random dropping techniques alone often yield unsatisfactory performance. \ding{186} Lastly, we observe that adopting initial connection and group normalization is universally effective across tens of classical graph datasets. Those findings urge more synergistic rethinking of those seminal works.

\vspace{-1em}
\section{Related Works}
\subsection{Graph Neural Networks and Training Deep GNNs}
\vspace{-0.1em}
GNNs ~\cite{zhou2018graph,kipf2016semi,chen2019equivalence,velivckovic2017graph,ying2018hierarchical,xu2018powerful,you2020graph,you2021graph,pmlr-v139-chen21p,hu2021vgai,zheng2021cold,zhou2021multi} have established state-of-the-art results on various tasks ~\cite{kipf2016semi,velivckovic2017graph,qu2019gmnn,verma2019graphmix,karimi2019explainable,you2020l2,pmlr-v119-you20a}. There are also numerous GNN variants~\cite{dwivedi2020benchmarking,scarselli2008graph,bruna2013spectral,kipf2016semi,hamilton2017inductive,battaglia2016interaction,monti2017geometric,velivckovic2018graph,xu2018how,morris2019weisfeiler,chen2019equivalence,murphy2019relational}. For example, Graph Convolutional Networks (GCNs) are widely adopted, which can be divided into spectral domain based methods~\cite{defferrard2016convolutional, kipf2016semi} and spatial domain bases methods~\cite{simonovsky2017dynamic, hamilton2017inductive}. Recent effort summarizes a series of useful tricks (e.g., label usage and loss function) to train shallow GNNs~\cite{wang2021bag}, which is different to the deep GNNs benchmark investigated in this paper.

Unlike other deep architectures, not every useful GNN has to be deep. For example, many graphs like social networks, are ``small-world" \cite{barcelo2019logical}, i.e., one node can reach any other node in a few hops. Hereby, adding more layers does not help model over those graphs since the receptive fields of just stacking a few layers would already suffice to provide global coverage. Many predictions in practice, such as in social networks, also mainly rely on short-range information from the local neighbourhood of a node. On the other hand, when the graph data is large in scale, does not have small-world properties, or its task requires long-range information to make prediction, then \textit{deeper GNNs do become necessary}. Examples include molecular graphs, as a molecule's chemical properties can depend on the atom combination at its opposite sides \cite{matlock2019deep}. The graphs with larger diameters (e.g., citation networks), point clouds or meshes also benefit from GNN depth to capture a whole object and its context \cite{li2019deepgcns,gong2020geometrically, zhou2021dirichlet}. 

With the prosperity of GNNs, understanding their training mechanism and limitation is of remarkable interest. As GNNs stack spatial aggregations recursively \cite{li2015gated, hamilton2017inductive}, the node representations will collapse to indistinguishable vectors~\cite{nt2019revisiting, oono2020graph, chen2020measuring}. Such over-smoothing phenomenon hinders the training of deep GNNs and the dependency modeling to high-order neighbors. Specifically, the over-smoothing usually arises when we naively increase the number of graph convolutional layers, which substantially decreases the discriminative power of node embeddings and ends up with a degraded performance on graph learning tasks like node classification and link prediction. Recently, there has been a series of techniques developed to relieve the over-smoothing issue, including \textit{skip connection}, \textit{graph normalization}, \textit{random dropping}. They will be detailed next.

\vspace{-1em}
\subsection{Skip Connection} 
Motivated by ResNets~\cite{he2016deep}, the skip connection was applied to GNNs to exploit node embeddings from the preceding layers, to relieve the over-smoothing issue. There are several skip connection patterns in deep GNNs, including: (1) residual connection connecting to the last layer~\cite{li2019deepgcns, li2018deeper}, (2) initial connection connecting to the initial layer~\cite{chen2020simple, klicpera2018predict, zhang2020revisiting}, (3) dense connection connecting to all the preceding layers~\cite{li2019deepgcns, li2018deeper, li2020deepergcn, luan2019break}, and (4) jumping connection combining all all the preceding layers only at the final graph convolutional layer~\cite{xu2018representation, liu2020towards}. Note that the last one of jumping connection is a simplified version of dense connection by omitting the complex skip connections at the intermediate layers.

\begin{table*}[t]
\caption{Configurations of basic hyperparameters adopted to implement different approaches for training deep GNNs on Cora~\cite{kipf2016semi}. Other graph datasets are refer to Section~\ref{sec:more_tech_details}.}
\vspace{-4mm}
\label{tab:settings_cora}
\centering
\resizebox{0.95\textwidth}{!}{
\begin{tabular}{@{}lrrrrr@{}}
\toprule
Methods & Total epoch & Learning rate \& Decay & Weight decay & Dropout & Hidden dimension \\
\midrule
Chen et al. (2020)~\cite{chen2020simple} & 100 & 0.01 & $5\times10^{-4}$ & 0.6 & 64 \\
Xu et al. (2018)~\cite{xu2018representation} & - & 0.005 & $5\times10^{-4}$ & 0.5 & \{16, 32\} \\
Klicpera et al. (2018)~\cite{klicpera2018predict} & 10000 & 0.01 & $5\times10^{-3}$ & 0.5 & 64 \\
Zhang et al. (2020)~\cite{zhang2020revisiting} & 1500 & \{0.001, 0.005, 0.01\} & - & \{0.1, 0.2, 0.3, 0.4, 0.5\} & 64 \\
Luan et al. (2019)~\cite{luan2019break} & 3000 & $1.66\times10^{-4}$ & $1.86\times10^{-2}$ & 0.65277 & 1024 \\
Liu et al. (2020)~\cite{liu2020towards} & 100 & 0.01 & 0.005 & 0.8 & 64 \\
Zhao et al. (2019)~\cite{zhao2019pairnorm} & 1500 & 0.005 & $5\times10^{-4}$ & 0.6 & 32 \\
Min et al. (2020)~\cite{scattering1} & 200 & 0.005 & 0 & 0.9 & - \\
Zhou et al. (2020)~\cite{zhou2020towards} & 1000 & 0.005 & $5\times10^{-4}$ & 0.6 & - \\
Zhou et al. (2020)~\cite{zhou2020understanding} & 50 & 0.005 & $1\times10^{-5}$  & 0 & - \\
Rong et al. (2020)~\cite{rong2020dropedge} & 400 & 0.01 & $5\times10^{-3}$ & 0.8 & 128 \\
Zou et al. (2020)~\cite{zou2019layer} & 100 & 0.001 & - & - & 256 \\
Hasanzadeh et al. (2020)~\cite{hasanzadeh2020bayesian} & 2000 & 0.005 & $5\times10^{-3}$ & 0 & 128 \\
\bottomrule
\end{tabular}}
\vspace{-1em}
\end{table*}

\begin{table*}[!ht]
\centering
\caption{The ``sweet point" hyperparameter configuration we used on representative datasets.}
\vspace{-4mm}
\resizebox{0.95\textwidth}{!}{
\begin{tabular}{@{}ccccc@{}}
\toprule
Settings & Cora & Citeseer & PubMed & OGBN-ArXiv\\
\midrule
\begin{tabular}[c]{@{}c@{}} \{Learning rate, Weight decay, \\ Dropout, Hidden dimension\} \end{tabular} & \{$0.005$, $5\times10^{-4}$, $0.6$, $64$\} & \{$0.005$, $5\times10^{-4}$, $0.6$, $256$\} & \{$0.01$, $5\times10^{-4}$, $0.5$, $256$\} & \{$0.005$, $0$, $0.1$, $256$\} \\
\bottomrule
\end{tabular}}
\vspace{-4mm}
\label{tab:experiment_settings}
\end{table*}

\vspace{-1em}
\subsection{Graph Normalization} A series of normalization techniques has been developed for deep GNNs, including batch normalization (BatchNorm)~\cite{ioffe2015batch}, pair normalization (PairNorm)~\cite{zhao2019pairnorm}, node normalization (NodeNorm)~\cite{zhou2020understanding}, mean normalization (MeanNorm)~\cite{yang2020revisiting}, differentiable group normalization (GroupNorm)~\cite{zhou2020towards}, and more. Their common mechanism is to re-scale node embeddings over an input graph to constrain pairwise node distance and thus alleviate over-smoothing. While BatchNorm and PairNorm normalize the whole input graph, GroupNorm first clusters node into groups and then normalizes each group independently. NodeNorm and MeanNorm operate on each node by re-scaling the associated node embedding with its standard deviation and mean values, respectively.

\vspace{-1em}
\subsection{Random Dropping} Dropout~\cite{srivastava2014dropout} refers to randomly dropping hidden units in a neural network with a pre-fixed probability, which effectively prevents over-fitting. Similar ideas have been inspired for GNNs. DropEdge~\cite{rong2020dropedge} and DropNode~\cite{dropedge2} have been proposed to randomly remove a certain number of edges or nodes from the input graph at each training epoch. Alternatively, they could be regarded as data augmentation methods, helping relieve both the over-fitting and over-smoothing issues in training very deep GNNs.

Besides the above widely-adopted tricks of skip connection, graph normalization, and random dropping, there exist some extra efforts made to optimize deep GNNs from other architectural perspectives. For example, the feature over-correlation is relieved to encourage deep GNNs to produce node embeddings with less redundant information~\cite{jin2022towards, guo2021orthogonal}. The grouped reversible graph connections and weight sharing among layers are leveraged to reduce memory complexity and train GNNs up to $1,000$ layers~\cite{li2021training}. Contrastive learning is used to generate the discriminative node representations and relieve the over-smoothing~\cite{zheng2021tackling}. Similar to SGC~\cite{wu2019simplifying} and DAGNN~\cite{liu2020towards}, DGMLP~\cite{zhang2021evaluating} has been proposed to establish deep GNNs by decoupling the graph convolutions and feature transformation.

\vspace{-1em}
\section{A Fair and Scalable Study of Tricks} \label{sec:tricks_study}


\subsection{Prerequisite: Unifying the Hyperparameter Configuration}
We carefully examine previous implementations of deep GNNs~\cite{chen2020simple,klicpera2018predict,liu2020towards,luan2019break, xu2018representation,scattering1,zhao2019pairnorm,zhou2020towards,rong2020dropedge,zhou2020understanding}, and list all their basic hyperparameters in Table~\ref{tab:settings_cora},~\ref{tab:settings_cite},~\ref{tab:settings_pubmed},~\ref{tab:settings_arxiv}. Those hyperparameters play significant roles in those methods' achievable performance, but their inconsistency challenges fair comparison of training techniques, which has been traditionally somehow overlooked in literature. 

We tune them all with an exhaustive grid search, and identify the most common and effective setting as in Table~\ref{tab:experiment_settings}, by choosing the configuration that performs the best across diverse backbones with different layers on the corresponding dataset. That configuration includes learning rate, weight decay, dropout rate, and the hidden dimension of multilayer perceptrons (MLP) in GNNs. We recommend it as a ``sweet point'' hyperparameter configuration, and strictly follow it in our experiments. 

For all experiments, deep GNNs are trained with a maximum of $1000$ epochs with an early stop patience of $100$ epochs. \textit{One hundred independent repetitions} are conducted for each experiment, and the average performances with the standard deviations of the node classification accuracies are reported in Table~\ref{tab:res_connection},~\ref{tab:res_connection_full},~\ref{tab:norm},~\ref{tab:norm_full},~\ref{tab:random_dropout},~\ref{tab:random_dropout_full},~\ref{tab:random_dropout_0.2}, and~\ref{tab:random_dropout_0.5}. We perform a comprehensive benchmark study of dozens of training approaches on four classical datasets, i.e., Cora, Citeseer, PubMed~\cite{kipf2016semi}, and OGBN-ArXiv~\cite{hu2020open} with $2/16/32$ layers GCN~\cite{kipf2016semi} and simple graph convolutional (SGC)~\cite{wu2019simplifying} backbones. GCN and SGC are chosen in our work because they are two most common backbone architectures for deep GNNs researches, following the standard in~\cite{li2019deepgcns,li2018deeper,klicpera2018predict,zhang2020revisiting,ioffe2015batch,ba2016layer,yang2020revisiting,zhou2020towards,xu2018representation, liu2020towards,srivastava2014dropout,rong2020dropedge,dropedge2,defferrard2016convolutional,li2020deepergcn,xu2021optimization,subgraphsample}. More details of these benchmark datasets and GNN backbones are collected in Appendix~\ref{sec:more_imp_details}. The Pytorch framework and Pytorch Geometric~\cite{Fey/Lenssen/2019} are used for all of our implementations. Specific setups for each group of training techniques are included in Section~\ref{sec:methods_connection},~\ref{sec:methods_norm}, and~\ref{sec:methods_dropping}.

\vspace{-1em}
\subsection{Skip Connection} \label{sec:methods_connection}

\begin{table*}[t]
\caption{Test accuracy (\%) under different skip connection mechanisms. Experiments are conducted on Cora, Citeseer, PubMed, and OGBN-ArXiv with $2/16/32$ layers GCN and SGC. Performance are averaged from $100$ independent repetitions, and their standard deviation are presented in Table~\ref{tab:res_connection_full}.}
\vspace{-4mm}
\label{tab:res_connection}
\centering
\resizebox{0.95\textwidth}{!}{
\begin{tabular}{@{}llcccccccccccc@{}}
\toprule
\multirow{2}{*}{Backbone} & \multirow{2}{*}{Settings} & \multicolumn{3}{c}{Cora} & \multicolumn{3}{c}{Citeseer} & \multicolumn{3}{c}{PubMed} & \multicolumn{3}{c}{OGBN-ArXiv}\\
\cmidrule(r){3-5} \cmidrule(r){6-8} \cmidrule(r){9-11} \cmidrule(r){12-14}
& & 2 & 16 & 32 & 2 & 16 & 32 & 2 & 16 & 32 & 2 & 16 & 32\\ 
\midrule
\multirow{5}{*}{GCN} & Residual & 74.73 & 20.05 & 19.57 & 66.83 & 20.77 & 20.90 & 75.27 & 38.84 & 38.74 & 70.19 & 69.34 & 65.09\\
& Initial & 79.00 & \textbf{78.61} & \textbf{78.74} & 70.15 & \textbf{68.41} & \textbf{68.36} & 77.92 & \textbf{77.52} & \textbf{78.18} & 70.16 & 70.50 & 70.23\\
& Jumping & 80.98 & 76.04 & 75.57 & 69.33 & 58.38 & 55.03 & 77.83 & 75.62 & 75.36 & \textbf{70.24} & \textbf{71.83} & \textbf{71.87}\\
& Dense & 77.86 & 69.61 & 67.26 & 66.18 & 49.33 & 41.48 & 72.53 & 69.91 & 62.99 & 70.08 & 71.29 & 70.94\\
& None & \textbf{82.38} & 21.49 & 21.22 & \textbf{71.46} & 19.59 & 20.29 & \textbf{79.76} & 39.14 & 38.77 & 69.46 & 67.96 & 45.48\\
\midrule
\multirow{5}{*}{SGC} & Residual & \textbf{81.77} & 82.55 & 80.14 & 71.68 & 71.31 & 71.00 & 78.87 & 79.86 & 79.07 & 69.09 & 66.52 & 61.83\\
& Initial & 81.40 & \textbf{83.66} & 83.77 & 71.60 & \textbf{72.16} & \textbf{72.25} & \textbf{79.11} & 79.73 & 79.74 & 68.93 & 69.24 & 69.15\\
& Jumping & 77.75 & 83.42 & \textbf{83.88} & 69.96 & 71.89 & 71.88 & 77.42 & \textbf{79.99} & \textbf{80.07} & 68.76 & 70.61 & 70.65\\
& Dense   & 77.31 & 81.24 & 77.66 & 70.99 & 67.75 & 66.35 & 77.12 & 72.77 & 74.84 & \textbf{69.39} & \textbf{71.42} & \textbf{71.52} \\
& None & 79.31 & 75.98 & 68.45 & \textbf{72.31} & 71.03 & 61.92 & 78.06 & 69.18 & 66.61 & 61.98 & 41.58 & 34.22\\
\bottomrule
\end{tabular}}
\vspace{-3mm}
\end{table*}

\textbf{Formulations.} Skip connections~\cite{li2019deepgcns,chen2020simple,li2018deeper,klicpera2018predict,zhang2020revisiting,xu2018representation,liu2020towards} alleviate the vanishing gradient and over-smoothing, and substantially improve the accuracy and stability of training deep GNNs. Let $\mathcal{G}=(\mathbf{A},\mathbf{X})$ denotes the graph data, where $\mathbf{A}\in\mathbb{R}^{n\times n}$ represents the adjacent matrix, $\mathbf{X}=[\mathbf{x}_1, \mathbf{x}_2,..., \mathbf{x}_n]^\top\in\mathbb{R}^{n\times d}$ is the corresponding input node features, $n$ is the number of nodes in the graph $\mathcal{G}$, and $d$ is the dimension of each node feature $\mathbf{x}_i$. In the adjacent matrix $\mathbf{A}$, the values of entries can be weighted, i.e., $\mathbf{A}_{i,j}\in(0,1)$. For a $L$-layer GNN, we can apply various types of skip connections after certain graph convolutional layers with the current and preceding embeddings $\mathbf{X}^{l}$, $0\le l\le L$. Four representative types of skip connections are investigated: 

\noindent $\star$ \textit{Residual Connection.}~\cite{li2019deepgcns, li2018deeper}

$\mathbf{X}^{l} = (1-\alpha)\cdot\mathbf{X}^{l}+\alpha\cdot\mathbf{X}^{l-1}$.

\noindent $\star$ \textit{Initial Connection.}~\cite{chen2020simple, klicpera2018predict, zhang2020revisiting} 

$\mathbf{X}^{l} = (1-\alpha)\cdot\mathbf{X}^{l}+\alpha\cdot\mathbf{X}^{0}$.

\noindent $\star$ \textit{Dense Connection.}~\cite{li2019deepgcns, li2018deeper, li2020deepergcn, luan2019break} 

$\mathbf{X}^{l} = \mathrm{COM}(\{\mathbf{X}^{k}, 0\le k \le l\})$.

\noindent $\star$ \textit{Jumping Connection.}~\cite{xu2018representation, liu2020towards} 

$\mathbf{X}^{L} = \mathrm{COM}(\{\mathbf{X}^{k}, 0\le k \le L\})$.

Here $\alpha$ in \textit{residual} and \textit{initial connections} is a hyperparameter to weigh the contributions of node features from the current layer $l$ and previous layers. In our case, the values of best performing $\alpha$ are searched from $\{0.1, 0.2, 0.4, 0.6, 0.8\}$ for each experiment. The initial embedding $\mathbf{X}^{0}$ is given by $\mathbf{X}$. \textit{Jumping connection} as a simplified case of \textit{dense connection} is only applied at the end of whole forward propagation process to combine the node features from all previous layers. 

We lastly introduce the combination functions $\mathrm{COM}$ used in dense and jumping connections: \ding{182} \textit{Concat}, $\mathbf{X}^{l} = \mathrm{MLP}([\mathbf{X}^{0},\mathbf{X}^{1}, \dots, \mathbf{X}^{l}])$; \ding{183} \textit{Maxpool}, $\mathbf{X}^{l} = max(stack([\mathbf{X}^{0},\mathbf{X}^{1}, \dots, \mathbf{X}^{l}]))$, the operation $max$ returns the maximum value for each dimension of a node feature vector; \ding{184} \textit{Attention}, $\mathbf{X}^{l} = squeeze(\mathbf{S}\hat{\mathrm{X}})$, $\mathbf{S}=\sigma(\mathrm{MLP}(\mathbf{X}))$, $\hat{\mathbf{X}}=stack([\mathbf{X}^{0},\mathbf{X}^{1}, \dots, \mathbf{X}^{l}])$. $\sigma(\cdot)$ is the activation function where $sigmoid$ is usually adopted~\cite{xu2018representation, liu2020towards}. $stack$ and $squeeze$ are utilized to rearrange the data dimension so that dimension can be matched throughout the computation. Note that, for every dense connection and jumping connection experiment, we examine all three options for $\mathrm{COM}$ in the set \{\textit{Concat}, \textit{Maxpool}, \textit{Attention}\} and report the one attaining the best performance.

\begin{figure}[h!]
\centering
\vspace{-1mm}
\includegraphics[width=0.91\linewidth]{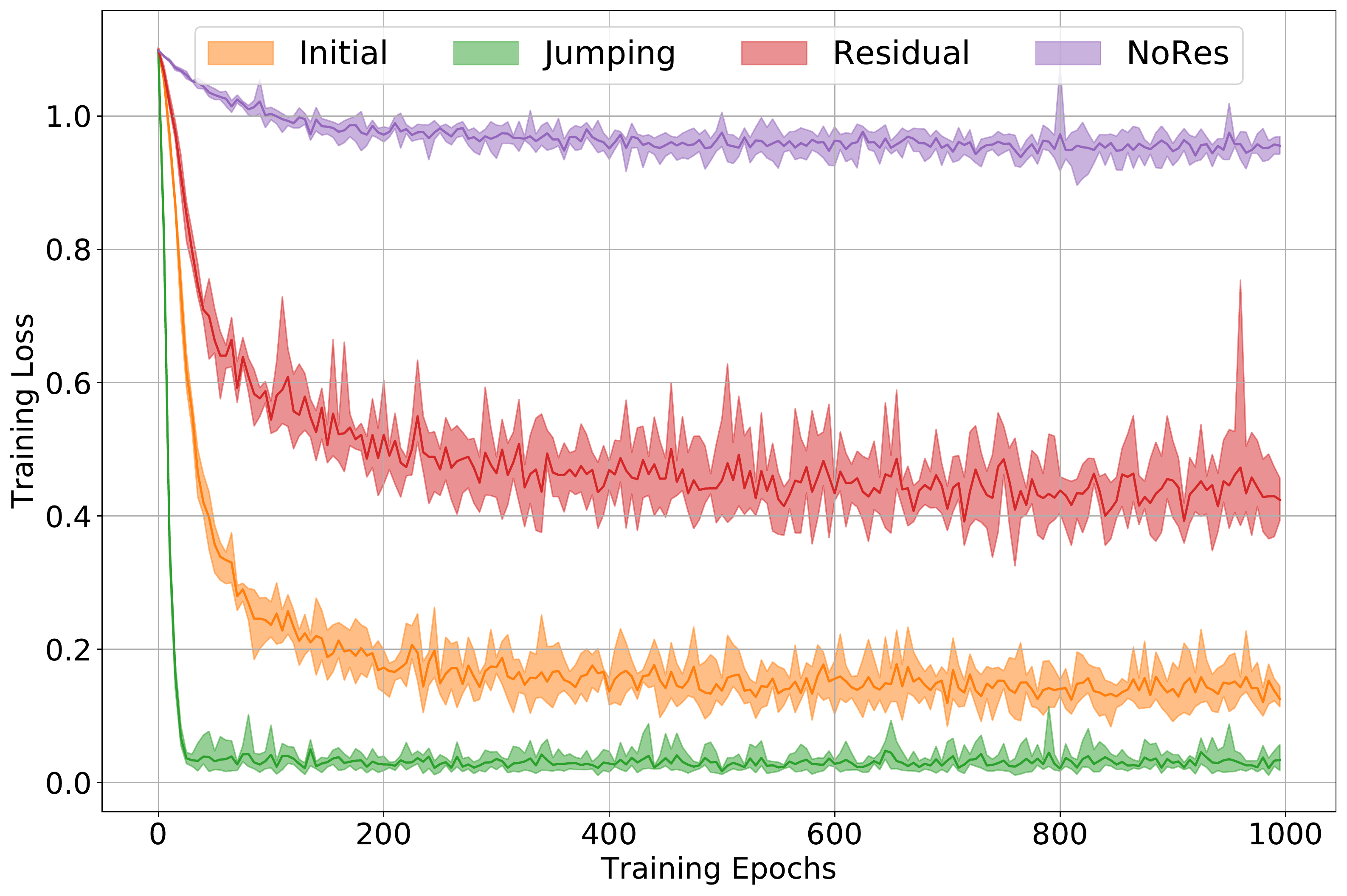}
\vspace{-5mm}
\caption{\small Training loss under different skip connections on PubMed with $16$ layers SGC. Dense connection is omitted due to different scales of loss magnitude. More are included in Section~\ref{sec:more_exp_results}.}
\label{fig:converge}
\vspace{-6mm}
\end{figure}

\textbf{Experimental observations.} From Table~\ref{tab:res_connection} and~\ref{tab:res_connection_full}, we can draw the following observations: $(i)$ In general, the initial and jumping connection perform relatively better than the others, especially for training very deep GNNs (i.e., $\ge 16$ layers). It demonstrates that these two skip connections more effectively mitigate the over-smoothing issue, assisting graph neural networks going deeper. $(ii)$ For shallow GCN backbones (e.g., $2$ layers), using skip connections can incur performance degradation, yet becomes beneficial on the large-scale OGBN-ArXiv dataset. Meanwhile, SGC backbones benefit from skip connections in most of the cases. $(iii)$ Residual connection only works for $2$-layer SGC on Cora. Although the residual connection also brings in shortcuts, its effects to overcome over-smoothing are diminished by increasing depth or growing dataset size, presumably due to its more ``local" information flow (only from the preceding one layer), compared to initial and dense connections whose information flows can be directly imported from much earlier layers. (iv) Although the dense connection on average brings significant accuracy improvements on OGBN-ArXiv with SGC, it sacrifices the training stability and leads to considerable performance variance, as consistently shown by Table~\ref{tab:res_connection_full}. $(v)$ Figure~\ref{fig:converge} reveals that skip connections substantially accelerate the training of deep GNNs, which is aligned with the analysis by concurrent work~\cite{xu2021optimization}.

\vspace{-1.1em}
\subsection{Graph Normalization} \label{sec:methods_norm}
\vspace{-0.5em}
\textbf{Formulations.} Graph normalization mechanisms  ~\cite{ioffe2015batch,ba2016layer,wu2018group,zhao2019pairnorm,zhou2020understanding,yang2020revisiting,zhou2020towards} are also designed to tackle over-smoothing. Nearly all graph normalization techniques are applied upon the intermediate node embedding matrix, after passing through several graph convolutions. We use $\bX^l=[\mathbf{x}^l_1, \mathbf{x}^l_2,..., \mathbf{x}^l_n]^\top\in\mR^{n\times d}$ to denote the node embedding matrix at layer $l$, where $\mathbf{x}^l_i\in\mR^{d\times 1}$ 
(we omit $l$ in later descriptions for notation's simplicity). Correspondingly, $\cx\in\mR^{n\times1}$ represents the $k$-th column of $\bX$, recording values of the $k$-th dimension of each feature embedding. 

Our investigated normalization mechanisms are formally depicted as follows:
\begin{table*}[t]
\caption{Test accuracy (\%) under different graph normalizations. Experiments are conducted on Cora, Citeseer, PubMed, and OGBN-ArXiv with $2/16/32$ layers GCN and SGC. Performance are averaged from $100$ independent repetitions, and their standard deviation are presented in Table~\ref{tab:norm_full}.}
\vspace{-4mm}
\label{tab:norm}
\centering
\resizebox{1\textwidth}{!}{
\begin{tabular}{@{}llcccccccccccc@{}}
\toprule
\multirow{2}{*}{Backbone} & \multirow{2}{*}{Settings} & \multicolumn{3}{c}{Cora} & \multicolumn{3}{c}{Citeseer} & \multicolumn{3}{c}{PubMed} & \multicolumn{3}{c}{OGBN-ArXiv}\\
\cmidrule(r){3-5} \cmidrule(r){6-8} \cmidrule(r){9-11} \cmidrule(r){12-14}
& & 2 & 16 & 32 & 2 & 16 & 32 & 2 & 16 & 32 & 2 & 16 & 32\\ 
\midrule
\multirow{5}{*}{GCN} 
 & BatchNorm & 69.91 & 61.20 & \textbf{29.05} & 46.27 & 26.25 & 21.82 & 67.15 & 58.00 & 55.98 & 70.44 & 70.52 & 68.74\\
& PairNorm & 74.43 & \textbf{55.75} & 17.67 & 63.26 & \textbf{27.45} & 20.67 & 75.67 & 71.30 & 61.54 & 65.74 & 65.37 & 63.32\\
 & NodeNorm & 79.87 & 21.46 & 21.48 & 68.96 & 18.81 & 19.03 & 78.14 & 40.92 & 40.93 & 70.62 & 70.75 & 29.94\\
 & MeanNorm & \textbf{82.49} & 13.51 & 13.03 & 70.86 & 16.09 & 7.70 & 78.68 & 18.92 & 18.00 & 69.54 & 70.40 & 56.94\\
 & GroupNorm & 82.41 & 41.76 & 27.20 & 71.30 & 26.77 & \textbf{25.82} & \textbf{79.78} & \textbf{70.86} & \textbf{63.91} & 69.70 & 70.50 & 68.14\\
 & CombNorm & 80.00 & 55.64 & 21.44 & 68.59 & 18.90 & 18.53 & 78.11 & 40.93 & 40.90  & \textbf{70.71} & \textbf{71.77} & \textbf{69.91}\\
 & NoNorm & 82.43 & 21.78 & 21.21 & \textbf{71.40} & 19.78 & 19.85 & 79.75 & 39.18 & 39.00 & 69.45 & 67.99 & 46.38\\
\midrule
\multirow{5}{*}{SGC}
 & BatchNorm & 79.32 & 15.86 & 14.40 & 61.60 & 17.34 & 17.82 & 76.34 & 54.22 & 29.49 & \textbf{68.58} & 65.54 & 62.33\\
 & PairNorm & 80.78 & 71.26 & 51.03 & 69.76 & 60.14 & 50.94 & 75.81 & 68.89 & 62.14 & 60.72 & 39.69 & 26.67\\
 & NodeNorm & 78.09 & \textbf{78.77} & 73.93 & 63.42 & 61.81 & 60.22 & 71.64 & 71.50 & 73.30 & 63.21 & 26.81 & 16.18\\
 & MeanNorm & 80.22 & 48.29 & 30.07 & 70.78 & 38.27 & 28.27 & 75.07 & 47.29 & 41.32 & 54.86 & 21.74 & 18.97\\
 & GroupNorm & \textbf{82.81} & 75.81 & \textbf{74.94} & 72.32 & 67.54 & 61.75 & \textbf{78.87} & \textbf{76.43} & \textbf{74.62} & 66.12 & \textbf{67.29} & \textbf{66.11}\\
 & CombNorm & 77.65 & 75.16 & 74.45 & 63.66& 59.97 & 54.52 & 71.67 & 71.50 & 72.23 & 65.73 & 54.37 & 47.52\\\
 & NoNorm & 79.38 & 75.93 & 68.75 & \textbf{72.36} & \textbf{71.06} & \textbf{62.64} & 78.01 & 69.06 & 66.55 & 61.96 & 41.43 & 34.24\\
\bottomrule
\end{tabular}}
\vspace{-4mm}
\end{table*}

\noindent $\star$ \textit{PairNorm.}~\cite{zhao2019pairnorm} 

\begin{small}$\tilx= \bx -\frac{1}{n}\sum_{i=1}^n \bx,\ \ \mathrm{PairNorm}(\bx;s)=\frac{s\cdot\tilx}{\sqrt{\frac{1}{n}\sum_{i=1}^n\|\tilx\|^2_2}}$
\end{small}.

\noindent $\star$ \textit{NodeNorm.}~\cite{zhou2020understanding} 

$\mathrm{NodeNorm}(\bx;p)=\frac{\bx}{\std(\bx)^\frac{1}{p}}$.

\noindent $\star$ \textit{MeanNorm.}~\cite{yang2020revisiting} 

$\mathrm{MeanNorm}(\cx) = \cx-\mE [\cx]$.

\noindent $\star$ \textit{BatchNorm.}~\cite{ioffe2015batch} 

$\mathrm{BatchNorm}(\cx)=\gamma\cdot\frac{\cx-\mE [\cx]}{\std(\cx)}+\beta$.

\noindent $\star$ \textit{GroupNorm.}~\cite{zhou2020towards} 

\begin{small}
$\mathrm{GroupNorm}(\bX;G,\lambda)=\bX+\lambda\cdot\sum_{g=1}^G\mathrm{BN}(\tilX_g),\tilX_g= \mathrm{softmax}(\bX\cdot\bU)[:,g] \circ \bX$
\end{small}.

Here $s$ in \textit{PairNorm} is a hyperparameter controlling the average pair-wise variance and we choose $s=1$ in our case. $p$ in \textit{NodeNorm} denotes the normalization order and our paper uses $p=2$. $std(\cdot)$ and $\mE[\cdot]$ are functions that calculate the standard deviation and mean value, respectively. $\circ$ in \textit{GroupNorm} is the operation of a row-wise multiplication, $\bU\in\mR^{d\times G}$ represents a learned transformation, $G$ means the number of groups, and $\lambda$ is the skip connection coefficient used in \textit{GroupNorm}. Specifically, in our implementation, we adopt $(G,\lambda)=(10,0.03)$ for Cora, $(5,0.01)$ for Pubmed, and $(10, 0.005)$ for Citeceer and OGBN-ArXiv. \textit{CombNorm} normalizes layer-wise graph embeddings by successively applying the two top-performing \textit{GroupNorm} and \textit{NodeNorm}.

\textbf{Experimental observations.} Extensive results are collected in Table~\ref{tab:norm} and~\ref{tab:norm_full}. We conclude in the following: ($i$) \textit{Dataset dependent observations.} On small datasets such as Cora and Citeceer, non-parametric normalization methods such as NodeNorm and PairNorm often present on-par performance with parametric GroupNorm, and GroupNorm displays better performance than CombNorm (i.e., GroupNorm + NodeNorm). On the contrary for large graphs such as OGBN-ArXiv, GroupNorm and CombNorm show a superior generalization ability than non-parametric normalizations, and adding NodeNorm on top of GroupNorm leads to extra improvements on deep GCN backbones. ($ii$) \textit{Backbone dependent observations.} We observe that even the same normalization technique usually appears diverse behaviors on different GNN backbones. For example, PairNorm is nearly always better than NodeNorm on $16$ and $32$ layers of GCN on three small datasets, and worse than NodeNorm on OGBN-ArXiv. However the case is almost reversed when we experiment with the SGC backbone, where NodeNorm is outperforming PairNorm on small graphs and inferior on OGBN-ArXiv, for $16/32$ layers GNNs. Meanwhile, we notice that SGC with normalization tricks achieve similar performance to GCN backbone on three small graphs, while consistently fall behind on large-scale OGBN graphs. ($iii$) \textit{Stability observations.} In general, training deep GNNs with normalization incurs more instability (e.g., large performance variance) along with the increasing depth of GNNs. BatchNorm demonstrates degraded performance with large variance on small graphs, and yet it is one of the most effective mechanism on the large graph with outstanding generalization and improved stability. ($iv$) Overall, across the five investigated norms, GroupNorm has witnessed most of top-performing scenarios, and PairNorm/NodeNorm occupied second best performance in GCN/SGC respectively, dependent on the scale of datasets. In contrast, MeanNorm performs the worst in training deep GCN/SGC, in terms of both achievable accuracy and performance stability.

\begin{table*}[t]
\caption{\small Test accuracy (\%) under different random dropping. Experiments are conducted on Cora, Citeseer, PubMed, and OGBN-ArXiv with $2/16/32$ layers GCN and SGC. Performance are averaged from $100$ repetitions, and standard deviations are reported in Table~\ref{tab:random_dropout_full},\ref{tab:random_dropout_0.2},\ref{tab:random_dropout_0.5}. Dropout rates are tuned for best performance.}
\vspace{-4mm}
\label{tab:random_dropout}
\centering
\resizebox{1\textwidth}{!}{
\begin{tabular}{@{}llcccccccccccc@{}}
\toprule
\multirow{2}{*}{Dataset} & \multirow{2}{*}{Settings} & \multicolumn{3}{c}{Cora} & \multicolumn{3}{c}{Citeseer} & \multicolumn{3}{c}{PubMed} & \multicolumn{3}{c}{OGBN-ArXiv} \\
\cmidrule(r){3-5} \cmidrule(r){6-8} \cmidrule(r){9-11} \cmidrule(r){12-14}
& & 2 & 16 & 32 & 2 & 16 & 32 & 2 & 16 & 32 & 2 & 16 & 32 \\ 
\midrule
\multirow{8}{*}{GCN} & No Dropout & 80.68 & \textbf{28.56} & \textbf{29.36} & 71.36 & 23.19 & \textbf{23.03} & 79.56 & 39.85 & 40.00 & \textbf{69.53} & 66.14 & 41.96 \\
& Dropout & \textbf{82.39} & 21.60 & 21.17 & \textbf{71.43} & 19.37 & 20.15 & \textbf{79.79} & 39.09 & 39.17 & 69.40 & 67.79 & 45.41 \\
& DropNode & 77.10 & 27.61 & 27.65 & 69.38 & 21.83 & 22.18 & 77.39 & 40.31 & 40.38 & 66.67 & 67.17 & 43.81 \\
& DropEdge & 79.16 & 28.00 & 27.87 & 70.26 & 22.92 & 22.92 & 78.58 & 40.61 & 40.50 & 68.67 & 66.50 & \textbf{51.70} \\
& LADIES & 77.12 & 28.07 & 27.54 & 68.87 & 22.52 & 22.60 & 78.31 & 40.07 & 40.11 & 66.43 & 62.05 & 40.41 \\
& DropNode+Dropout & 81.02 & 22.24 & 18.81 & 70.59 & 24.49 & 18.23 & 78.85 & 40.44 & 40.37 & 68.66 & 68.27 & 44.18 \\
& DropEdge+Dropout & 79.71 & 20.45 & 21.10 & 69.64 & 19.77 & 18.49 & 77.77 & 40.71 & 40.51 & 66.55 & \textbf{68.81} & 49.82 \\
& LADIES+Dropout & 78.88 & 19.49 & 16.92 & 69.02 & \textbf{27.17} & 18.54 & 78.53 & \textbf{41.43} & \textbf{40.70} & 66.35 & 65.13 & 39.99 \\
\midrule
\multirow{8}{*}{SGC} & No Dropout & 77.55 & 73.99 & 66.80 & 71.80 & \textbf{72.69} & 70.50 & 77.59 & 69.74 & 67.81 & \textbf{62.34} & \textbf{42.54} & \textbf{34.76} \\
& Dropout & 79.37 & 75.91 & 68.40 & \textbf{72.35} & 71.21 & 62.35 & 78.04 & 69.12 & 66.53 & 61.96 & 41.47 & 34.22 \\
& DropNode & 78.57 & 76.99 & \textbf{72.93} & 71.87 & 72.50 & \textbf{70.60} & 77.63 & 72.51 & 68.16 & 61.21 & 40.52 & 34.64 \\
& DropEdge & 78.68 & 70.65 & 44.00 & 71.94 & 69.43 & 45.13 & \textbf{78.26} & 68.39 & 52.08 & 62.06 & 41.03 & 33.61 \\
& LADIES & 78.50 & \textbf{78.35} & 72.71 & 71.88 & 71.69 & 69.80 & 77.65 & \textbf{74.86} & \textbf{72.27} & 61.49 & 38.96 & 33.17\\ 
& DropNode+Dropout & \textbf{80.60} & 74.83 & 55.04 & 72.33 & 70.30 & 65.85 & 78.10 & 67.98 & 52.01 & 61.54 & 39.48 & 32.63 \\
& DropEdge+Dropout & 80.27 & 76.19 & 66.08 & 72.09 & 66.48 & 35.55 & 77.63 & 69.65 & 67.55 & 60.21 & 39.12 & 33.81 \\
& LADIES+Dropout & 79.81 & 74.72 & 66.62 & 71.85 & 69.24 & 50.81 & 77.46 & 70.54 & 67.94 & 60.27 & 31.41 & 24.86 \\
\bottomrule
\end{tabular}}
\vspace{-4mm}
\end{table*}

\vspace{-1em}
\subsection{Random Dropping} \label{sec:methods_dropping}

\textbf{Formulations.} Random dropping~\cite{srivastava2014dropout,rong2020dropedge,dropedge2,NEURIPS2019_91ba4a44} is another group of approaches to address the over-smoothing issue by randomly removing or sampling a certain number of edges or nodes at each training epoch. Theoretically, it is effective in decelerating the over-smoothing and relieving the information loss caused by dimension collapse~\cite{rong2020dropedge,dropedge2}. 

As described above, $\mathbf{X}^{l} \in \mathbb{R}^{n \times d}$ is the node embeddings at the $l$-th layer. Let $\Re(\mathbf{A}) = (\mathbf{I} + \mathbf{D})^{-1/2}(\mathbf{I} +\mathbf{A})(\mathbf{I}+\mathbf{D})^{-1/2}$ denotes a normalization operator, where $\mathbf{A} \in \mathbb{R}^{n \times n}$ is the adjacency matrix. Then, a vanilla GCN layer can be written as $\mathbf{X}^{l+1} = \Re(\mathbf{A}) \mathbf{X}^{l} \mathbf{W}^{l}$, where $\mathbf{W}^{l} \in \mathbb{R}^{d \times d}$ denotes the weight matrix of the $l$-th layer.

$\star$ \textit{Dropout}.~\cite{srivastava2014dropout} $\mathbf{X}^{l+1} = \Re(\mathbf{A}) (\mathbf{X}^{l} \odot \mathbf{Z}^{l}) \mathbf{W}^{l}$, where $\mathbf{Z}^{l} \in \{0, 1\}^{n 
\times d}$ is a binary random matrix, and each element $\mathbf{Z}^{l}_{ij} \sim \operatorname{Bernoulli}(\pi)$ is drawn from Bernoulli distribution.

$\star$ \textit{DropEdge}.~\cite{rong2020dropedge} $\mathbf{X}^{l+1} = \Re(\mathbf{A} \odot \mathbf{Z}^{l}) \mathbf{X}^{l} \mathbf{W}^{l}$, where $\mathbf{Z}^{l} \in \{0, 1\}^{n \times n}$ is a binary random matrix, and each element $\mathbf{Z}^{l}_{ij} \sim \operatorname{Bernoulli}(\pi)$ is drawn from Bernoulli distribution.

$\star$ \textit{DropNode}.~\cite{dropedge2} $\mathbf{X}^{l+1} = \Re(\mathbf{Q}^{l+1} \mathbf{A} \operatorname{diag}(\mathbf{Z}^{l}) \mathbf{Q}^{l\top}) \mathbf{X}^{l} \mathbf{W}^{l}$, where $\mathbf{Z}^{l} \in \{0, 1\}^{n}$ is a binary random matrix, and each element $\mathbf{Z}^{l}_{j} \sim \operatorname{Bernoulli}(\pi)$ is drawn from Bernoulli distribution. Suppose the index of selected nodes are $i_1, i_2, \cdots, i_{s_l}$, $\mathbf{Q}^{l} \in \{0,1\}^{s_l \times n}$ is a row selection matrix, where $\mathbf{Q}_{k,m}^l = 1$ if and only if the $m = i_k$, and $1\le k \le s_l$, $1\le m \le n$.

$\star$ \textit{LADIES}.~\cite{NEURIPS2019_91ba4a44} $\mathbf{X}^{l+1} = \Re(\mathbf{Q}^{l+1} \mathbf{A} \operatorname{diag}(\mathbf{Z}^{l}) \mathbf{Q}^{l\top}) \mathbf{X}^{l} \mathbf{W}^{l}$, where $\mathbf{Z}^{l} \in \{0, 1\}^{n}$ is a binary random matrix with $\lVert \mathbf{Z}^{l} \rVert_0 = s_l$, and each element $\mathbf{Z}^{l}_{j} \sim \operatorname{Bernoulli}\left( \frac{\lVert \mathbf{Q}^{l} \mathbf{A}_{:,j} \rVert_2^2}{\lVert \mathbf{Q}^{l} \mathbf{A} \rVert_F^2} \right)$ is drawn from Bernoulli distribution. $\mathbf{Q}^{l} \in \{0,1\}^{s_l \times n}$ is the row selection matrix sharing the same definition in \textit{DropNode}.

For these dropout techniques, the neuron dropout rate follows the ``sweet point" configuration in Table~\ref{tab:experiment_settings}, and the graph dropout rates are picked from $\{0.2, 0.5, 0.7\}$ to attain best results. All presented results adopt the layer-wise dropout scheme consistent with the one in~\cite{zou2019layer}.

\textbf{Experimental observations.} Main evaluation results of random dropping mechanisms are presented in Table~\ref{tab:random_dropout} and~\ref{tab:random_dropout_full}, and additional results are referred to Table~\ref{tab:random_dropout_0.2} and~\ref{tab:random_dropout_0.5}. Several interesting findings can be summarized as follows: ($i$) The overall performance of random dropping is not as significant as skip connection and graph normalization. Such random dropping tricks at most retard the performance drop, but can not boost deep GNNs to be stronger than shallow GNNs. For examples, on Cora dataset, all dropouts fail to improve deep GCNs; on OGBN-ArXiv dataset, all dropouts also fail to benefits deep SGCs. On other graph datasets, only a small subset of dropping approaches can relieve performance degradation, while others tricks still suffer accuracy deterioration. ($ii$) The effects of random dropping vary drastically across models, datasets, and layer numbers. It is hard to find a consistently improving technique. To be specific, we observe that, \textit{Sensitive to models:} DropNode on SGC has been shown as an effective training trick for Cora dataset, but turns out to be useless when cooperating with GCN;
\textit{Sensitive to datasets:} On Pubmed dataset, the LADIES is the best trick, while it becomes the worst one on OGBN-ArXiv; \textit{Sensitive to layer number:} Deepening SGC from 16 layers to 32 layers, the performance of DropEdge declines significantly from $65\%\sim70\%$ to $40\%\sim50\%$.
A noteworthy exception is LADIES trick on PubMed, which consistently achieves the best result for both $16/32$-layer GCN and SGC.
($iii$) Dropout (with common rate setting $0.6$) is demonstrated to be beneficial and stronger than node or edge dropping for shallow GCNs (i.e., 2 layers). However, this dropout rate downgrades the performance when GCN goes deeper. Combining Dropout with graph dropping may not always be helpful. For instance, incorporating dropout into LADIES downgrades the accuracy of 32 layers SGC from $69.8\%$ to $50.8\%$.
($iv$) SGC with random dropping shows better tolerance for deep architectures, and more techniques presented in Table~\ref{tab:random_dropout} can improve the generalization ability compared to the plain GCN. Meanwhile, with the same random dropping, SGC steadily surpasses GCN by a substantial performance margin for training deep GNN (e.g., $\ge 16$ layers) on three small graph datasets, while GCN shows a superior accuracy on large-scale OGBN graphs.
($v$) From the stability perspective, we notice that SGC consistently has smaller performance variance than GCN of which instability is significantly enlarged by the increasing depth.
\vspace{-1em}
\begin{table}[!ht]
\caption{\small Average test accuracy (\%) and its standard deviations from $100$ independent repetitions w. or w.o. the identity mapping.}
\vspace{-4mm}
\label{tab:other_trick}
\centering
\resizebox{0.9\linewidth}{!}{
\begin{tabular}{@{}lcccc@{}}
\toprule
\multirow{2}{*}{Dataset} & \multirow{2}{*}{\begin{tabular}[c]{@{}c@{}} \textit{Identity} \\ \textit{Mapping} \end{tabular}} & \multicolumn{3}{c}{GCNs} \\
\cmidrule(r){3-5} 
& & 2 & 16 & 32 \\ 
\midrule
\multirow{2}{*}{Cora} & with & \textbf{82.98$\pm$0.75} & \textbf{67.23$\pm$3.61} & \textbf{40.57$\pm$11.7} \\ 
& without & 82.38$\pm$0.33 & 21.49$\pm$3.84 & 21.22$\pm$3.71 \\ 
\midrule
\multirow{2}{*}{Citeseer} & with & 68.25$\pm$0.71 & \textbf{56.39$\pm$4.36} & \textbf{35.28$\pm$5.85} \\
& without & \textbf{71.46$\pm$0.44} & 19.59$\pm$1.96 & 20.29$\pm$1.79 \\
\midrule
\multirow{2}{*}{PubMed} & with & 79.09$\pm$0.48 & \textbf{79.55$\pm$0.41} & \textbf{73.74$\pm$0.71} \\
& without & \textbf{79.76$\pm$0.39} & 39.14$\pm$1.38 & 38.77$\pm$1.20 \\
\midrule
\multirow{2}{*}{OGBN-ArXiv} & with & \textbf{71.08$\pm$0.28} & \textbf{69.22$\pm$0.84} & \textbf{62.85$\pm$2.22}\\
& without & 69.46$\pm$0.22 & 67.96$\pm$0.38 & 45.48$\pm$4.50 \\
\bottomrule
\end{tabular}}
\vspace{-2mm}
\end{table}

\begin{figure*}[t]
    \centering
    \includegraphics[width=0.9\linewidth]{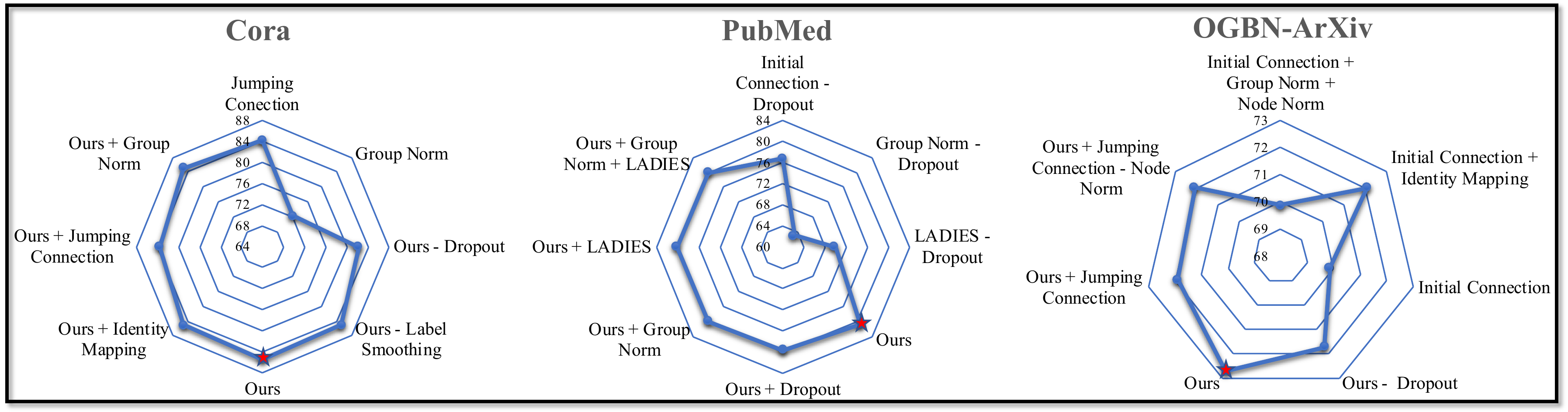}
    \vspace{-4mm}
    \caption{\small Average test accuracy (\%) from $100$ independent repetitions on Cora, PubMed, and OGBN-ArXiv graphs respectively. \textit{Ours} indicates our best combo of tricks on the corresponding dataset. $-/+$ means add/remove certain tricks from the trick combos. The top-performing setups are highlighted with \textcolor{red}{red} points.}
    \label{fig:abaltion}
    \vspace{-4mm}
\end{figure*}

\begin{table*}[t]
\caption{\small Test accuracy (\%) comparison with other previous state-of-the-art frameworks. Experiments are conducted on Cora, Citeseer, PubMed, and OGBN-ArXiv with $2/16/32$ GNNs. Performance are averaged from $100$ repetitions, and standard deviations are reported in Table~\ref{tab:model_comparison_full}. The superior performance achieved by \textbf{our best tricks combo} with deep GCNs (i.e., $\ge 16$ layers) are highlighted in the first raw.}
\vspace{-4mm}
\label{tab:model_comparison}
\centering
\resizebox{0.9\textwidth}{!}{
\begin{tabular}{@{}lcccccccccccc@{}}
\toprule
\multirow{2}{*}{Model}  & \multicolumn{3}{c}{Cora (\textbf{Ours: 85.48})} & \multicolumn{3}{c}{Citeseer (\textbf{Ours: 73.35})} & \multicolumn{3}{c}{PubMed (\textbf{Ours: 80.76})} & \multicolumn{3}{c}{ArXiv (\textbf{Ours: 72.70})}\\
\cmidrule(r){2-4} \cmidrule(r){5-7} \cmidrule(r){8-10} \cmidrule(r){11-13}
& 2 & 16 & 32 & 2 & 16 & 32 & 2 & 16 & 32 & 2 & 16 & 32\\ 
\midrule
SGC~\cite{wu2019simplifying} & 79.31 & 75.98 & 68.45 & 72.31 & 71.03 & 61.92 & 78.06 & 69.18 & 66.61 & 61.98 & 41.58 & 34.22 \\
DAGNN~\cite{liu2020towards} & 80.30 & 84.14 & 83.39 & 18.22 & 73.05 & 72.59 & 77.74 & 80.32 & 80.58 & 67.65 & 71.82 & 71.46  \\
GCNII~\cite{chen2020simple} & 82.19 & 84.69 & 85.29 & 67.81 & 72.97 & 73.24 & 78.05 & 80.03 & 79.91 & 71.24 & 72.61 & 72.60 \\
JKNet~\cite{xu2018representation} & 79.06 & 72.97 & 73.23 & 66.98 & 54.33 & 50.68 & 77.24 & 64.37 & 63.77 & 63.73 & 66.41 & 66.31  \\
APPNP~\cite{klicpera2018predict} & 82.06 & 83.64 & 83.68 & 71.67 & 72.13 & 72.13 & 79.46 & 80.30 & 80.24 & 65.31 & 66.95 & 66.94  \\
GPRGNN~\cite{chien2021adaptive} & 82.53 & 83.69 & 83.13 & 70.49 & 71.39 & 71.01 & 78.73 & 78.78 & 78.46 & 69.31 & 70.30 & 70.18 \\
\bottomrule
\end{tabular}}
\vspace{-4mm}
\end{table*}

\vspace{-1.5em}
\subsection{Other Tricks}
Although there are multiple other tricks in literature, such as shallow sub-graph sampling~\cite{subgraphsample} and geometric scattering transform of adjacency matrix~\cite{scattering1}, most of them are intractable to be incorporated into the existing deep GNN frameworks. To be specific, the shallow sub-graph sampling has to sample the shallow neighborhood for each node iteratively and the geometric scattering requires multiscale wavelet transforms on matrix $\mathbf{A}$. Both of them do not directly propagate message based upon $\mathbf{A}$, which is deviate from the common fashion of deep GNNs. Fortunately, we find that the \textit{Identity Mapping} proposed in~\cite{chen2020simple} is effective to augment existing frameworks and relieves the over-smoothing and overfitting issues. It can be depicted as follows:

$\star$ \textit{Identity Mapping}.~\cite{chen2020simple} $\mathbf{X}^{l+1} = \sigma(\mathbf{A}\mathbf{X}^{l} (\beta_l\cdot\mathbf{W}^{l} + (1-\beta_l)\cdot\mathbf{I}))$, $\beta_l$ is computed by $\beta_l = \log(\frac{\lambda}{l} + 1)$, where $\lambda$ is a positive hyperparameter, and $\beta_l$ decreases with layers to avoid the overfitting. In our case, $\lambda$ is grid searched for different graph datasets: $\lambda=0.1$ for Cora, Citeser and PubMed, and $\lambda=0.5$ for OGBN-ArXiv. 

\textbf{Experimental observations.} As shown in Table~\ref{tab:other_trick}, we conduct experiments on $2/16/32$ layers GCNs with or without the identity mapping. From the results on \{Cora, Citeseer, PubMed, OGBN-ArXiv\} datasets, ($i$) we find the identity mapping consistently brings deep GCNs (e.g., $16/32$ layers) significant accuracy improvements up to \{$45.74\%$, $36
.80\%$, $40.41\%$, $17.37\%$\}, respectively. It empirically evidenced the effectiveness of identity mapping technique in mitigating over-smoothing issue, particularly for small graphs. However, on shallow GCNs (e.g., $2$ layers), identity mapping does not work very well, which either obtains a marginal gain or even incurs some degradation. ($ii$) Although the generalization for deep GCNs on small-scale graphs (e.g., Cora and Citeseer) is largely enhanced, it also amplify the performance instability. 

\begin{table*}[t]
\caption{\small Transfer studies of our trick combos based on SGC (\textit{Ours}). Comparison are conducted on eight hold out datasets with two widely adopted baselines, i.e., GCNII and DAGNN. Performance are averaged from $100$ independent runs. All GNNs have $32$ layers.}
\vspace{-4mm}
\centering
\resizebox{0.95\textwidth}{!}{
\begin{tabular}{@{}lccccccccccccccccccccccccc@{}}
\toprule
Category & CS  & Physics & Computers & Photo & Texas &  Wisconsin &  Cornell &  Actor  \\ \midrule
SGC & 70.52$\pm$3.96 & 91.46$\pm$0.48 & 37.53$\pm$0.20 & 26.60$\pm$4.64 & 56.41$\pm$4.25 & 51.29$\pm$6.44 & 58.57$\pm$3.44 & 26.17$\pm$1.15\\
DAGNN & 89.60$\pm$0.71 & 93.31$\pm$0.60 & \textbf{79.73$\pm$3.63} & 89.96$\pm$1.16 & 57.68$\pm$5.07 & 50.84$\pm$6.62 & 58.43$\pm$3.93 & 27.73$\pm$1.08\\
GCNII & 71.67$\pm$2.68 & 93.15$\pm$0.92 & 37.56$\pm$0.43 & 62.95$\pm$9.41 & 69.19$\pm$6.56 & 70.31$\pm$4.75 & 74.16$\pm$6.48 & 34.28$\pm$1.12\\
JKNet & 81.82$\pm$3.32 & 90.92$\pm$1.61 & 67.99$\pm$5.07 & 78.42$\pm$6.95 & 61.08$\pm$6.23 & 52.76$\pm$5.69 & 57.30$\pm$4.95 & 28.80$\pm$0.97\\
APPNP & \textbf{91.61$\pm$0.49} & 93.75$\pm$0.61 & 43.02$\pm$10.16 & 59.62$\pm$23.27 & 60.68$\pm$4.50 & 54.24$\pm$5.94 & 58.43$\pm$3.74 & 28.65$\pm$1.28 \\
GPRGNN & 89.56$\pm$0.47 & 93.49$\pm$0.59 & 41.94$\pm$9.95 & \textbf{91.74$\pm$0.81} & 62.27$\pm$4.97 & 71.35$\pm$5.56 & 58.27$\pm$3.96 & 29.88$\pm$1.82\\
DGMLP~\cite{zhang2021evaluating} & 88.13$\pm$1.06 & 92.07$\pm$0.31 & 37.4$\pm$0.00 & 25.31$\pm$0.00 & 64.86$\pm$0.00 & 52.94$\pm$0.0 & 54.05$\pm$0.00 & 25.45$\pm$0.15 \\
\midrule
Ours on SGC & 90.74$\pm$0.66 & \textbf{94.12$\pm$0.56} & 75.40$\pm$1.88 & 88.29$\pm$1.73 & \textbf{79.68$\pm$3.77} & \textbf{83.59$\pm$3.32} & \textbf{81.24$\pm$5.77} & \textbf{36.84$\pm$0.70}\\
\bottomrule
\end{tabular}}
\vspace{-3mm}
\label{tab:compare_other_data}
\end{table*}

\begin{figure*}[t]
    \centering
    \includegraphics[width=0.95\textwidth]{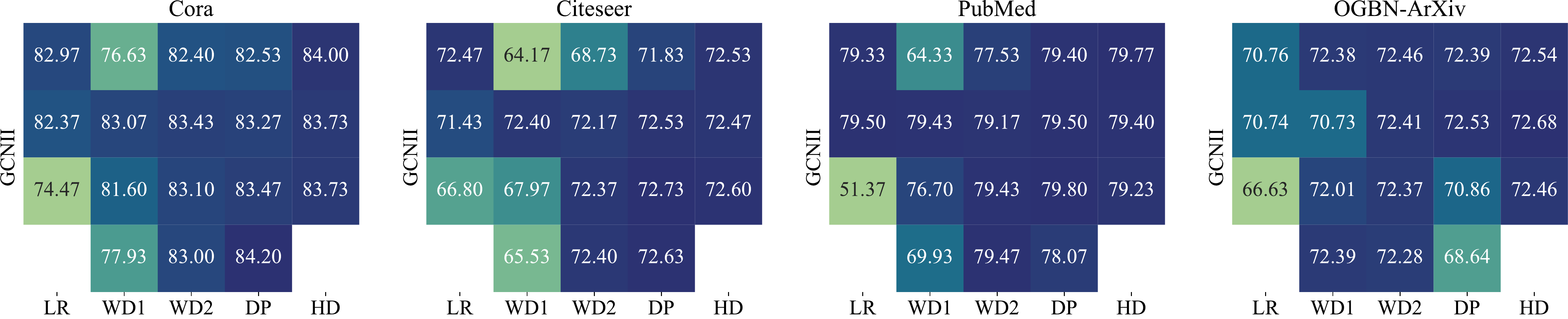}
    \vspace{-4mm}
    \caption{The hyperparameter investigation of GCNII on Cora, Citeseer, Pubmed, and OGBN-ArXiv. Node classification accuracies are presented in a manner of heatmap. From top to bottom, learning rate (LR), weight decay for the hidden layers (WD 1), weight decay for the prediction layer (WD 2), dropout (DP), and hidden dimension (HD) traverse from left to right in $\{0.01, 0.001, 0.0001\}$, $\{0, 0.01, 0.001, 0.0001\}$, $\{0, 1\times10^{-4}, 2\times10^{-4}, 5\times10^{-4}\}$, $\{0.1, 0.2, 0.5, 0.7\}$, and $\{128, 256, 512\}$, respectively.}
    \label{fig:hp}
    \vspace{-3mm}
\end{figure*}

\begin{table*}[t]
\caption{\small Training time per epoch (ms) and the max allocated memory (MB) for skip connection tricks.}
\vspace{-4mm}
\label{tab:complex_skip}
\centering
\resizebox{1\textwidth}{!}{
\begin{tabular}{@{}llcccccccccccc@{}}
\toprule
\multirow{3}{*}{Backbone} & \multirow{3}{*}{Settings} & \multicolumn{6}{c}{Cora} & \multicolumn{6}{c}{Citeseer} \\
\cmidrule(r){3-8} \cmidrule(r){9-14} 
& & \multicolumn{2}{c}{2} & \multicolumn{2}{c}{16} & \multicolumn{2}{c}{32} & \multicolumn{2}{c}{2} & \multicolumn{2}{c}{16} & \multicolumn{2}{c}{32} \\
\cmidrule(r){3-4} \cmidrule(r){5-6} \cmidrule(r){7-8} \cmidrule(r){9-10} \cmidrule(r){11-12} \cmidrule(r){13-14}
& & Time & Memory & Time & Memory & Time & Memory & Time & Memory & Time & Memory & Time & Memory \\
\midrule
\multirow{5}{*}{GCN} & None &3.09 & 191.92 &20.12&196.35 &30.36&201.40 &5.49&511.80 &20.54&529.13 &33.99&548.94 \\
& Residual &6.03&43.32 &23.56&77.83 &36.78&117.26 &6.07&161.10 &24.81&326.90 &39.71&515.66 \\
& Initial &6.13&43.32 &23.11&77.83 &36.22&117.26 &6.39&161.10 &25.11&325.27 &36.48&514.03  \\
& Jumping &5.37&43.32 &22.13&84.96 &32.32&134.45 &5.40&161.11 &24.25&365.55 &35.18&606.49 \\
& Dense &6.94&44.80 &32.39&180.97 &40.78&502.90 &7.17&168.84 &33.11&895.55 &97.50&2623.16 \\
\midrule
\multirow{5}{*}{SGC} & None &4.23&42.56 &10.95&67.16 &19.24&95.27 &3.96&156.99 &10.99&265.31 &18.12&390.08 \\
& Residual &4.66&42.56 &16.04&67.16 &23.06&95.27 &4.44&155.09 &15.43&264.26 &23.49&389.04 \\
& Initial &4.53&42.56 &15.05&67.16 &23.19&95.27 &4.88&156.99 &16.19&264.26 &23.51&389.04 \\
& Jumping &4.06&42.57 &13.14&73.63 &20.40&111.80 &3.96&157.00 &13.61&301.56 &20.30&478.52 \\
& Dense &5.77&44.04 &25.61&169.64 &39.11&480.24 &5.88&164.09 &30.31&830.34 &89.48&2493.59 \\
\toprule
\multirow{1}{*}{Backbone} & \multirow{1}{*}{Settings} & \multicolumn{6}{c}{PubMed} & \multicolumn{6}{c}{OGBN-ArXiv} \\ \midrule
\multirow{5}{*}{GCN} & None &8.36&535.83 &64.80&1298.71 &130.28&2350.06 &133.14&5856.61 &1136.12&14058.69 &2295.50&23432.50 \\
& Residual &11.99&445.36 &76.30&1365.64 &150.39&2415.98 &162.19&6399.08 &1226.54&14601.16 &2447.16&23974.96 \\
& Initial &12.08&445.36 &76.30&1365.64 &150.21&2415.98 &162.42&6399.08 &1227.36&14601.16 &2459.27&23974.96 \\
& Jumping &12.29&445.36 &74.00&1525.13 &145.12&2883.86 &160.46&6399.15 &1209.07&14602.33 &2433.02&24979.94 \\
& Dense &15.07&486.37 &186.55&4240.60 &575.00&13360.93 &186.89&6732.33 &2202.83&37004.45 & 4001.04&117163.48\\
\midrule
\multirow{5}{*}{SGC} & None &8.99&424.60 &49.02&1066.76 &95.34&1800.10 &135.20&6232.20 &1004.12&12109.80 &2014.96&18826.87 \\
& Residual &10.10&424.60 &57.31&1066.66 &111.84&1796.37 &144.02&6232.20 &1073.04&12108.55 &2137.83&18824.99 \\
& Initial &9.88&424.60 &57.29&1066.66 &111.81&1796.37 &143.88&6232.20 &1074.09&12108.55 &2149.85&18824.99 \\
& Jumping &9.29&424.61 &54.86&1207.24 &106.82&2248.57 &142.14&6232.28 &1058.32&12110.97 &2125.41&19666.49 \\
& Dense &12.68&465.62 &166.10&3917.57 &531.66&12721.31 &168.23&6565.46 &2037.07&34512.94 & 1935.8&53336.16\\
\bottomrule
\end{tabular}}
\vspace{-2mm}
\end{table*}

\begin{table*}[!ht]
\caption{\small Training time per epoch (ms) and the max allocated memory (MB) for normalization tricks.}
\vspace{-4mm}
\label{tab:complex_norm}
\centering
\resizebox{1\textwidth}{!}{
\begin{tabular}{@{}llcccccccccccc@{}}
\toprule
\multirow{3}{*}{Backbone} & \multirow{3}{*}{Settings} & \multicolumn{6}{c}{Cora} & \multicolumn{6}{c}{Citeseer} \\
\cmidrule(r){3-8} \cmidrule(r){9-14} 
& & \multicolumn{2}{c}{2} & \multicolumn{2}{c}{16} & \multicolumn{2}{c}{32} & \multicolumn{2}{c}{2} & \multicolumn{2}{c}{16} & \multicolumn{2}{c}{32} \\
\cmidrule(r){3-4} \cmidrule(r){5-6} \cmidrule(r){7-8} \cmidrule(r){9-10} \cmidrule(r){11-12} \cmidrule(r){13-14}
& & Time & Memory & Time & Memory & Time & Memory & Time & Memory & Time & Memory & Time & Memory \\
\midrule
\multirow{6}{*}{GCN} & None &3.09&191.92 &20.12&196.35 &30.36&201.40 &5.49&511.80 &20.54&529.13 &33.99&548.94 \\
& BatchNorm &3.57&191.93 &23.15&196.40 &35.97&201.51 &5.84&511.81 &24.58&529.23 &38.17&606.99 \\
& PairNorm &5.90&191.92 &35.19&196.35 &41.79&201.40 &6.76&511.80 &36.61&529.13 &42.99&607.39 \\
& NodeNorm &5.26&191.92 &32.45&196.35 &41.91&201.40 &6.57&511.80 &33.79&529.13 &38.97&707.63\\
& MeanNorm &3.82&191.92 &25.56&196.35 &38.25&201.40 &5.97&511.80 &26.53&529.13 &37.82&548.94 \\
& GroupNorm &12.92&191.95 &40.62&196.67 &59.52&348.51 &13.20&511.89 &40.27&858.86 &78.45&1625.21 \\
& CombNorm &14.46&191.95 &46.57&208.48 &68.62&389.83 &14.24&511.89 &44.06&958.53 &86.17&1827.35 \\
\midrule
\multirow{6}{*}{SGC} & None &4.23&42.56 &10.95&67.16 &19.24&95.27 &3.96&156.99 &10.99&265.31 &18.12&390.08 \\
& BatchNorm &4.45&43.23 &17.78&77.15 &26.09&115.91 &5.05&160.26 &17.80&315.04 &28.38&491.93  \\
& PairNorm &7.28&43.23 &28.32&77.08 &38.63&115.78 &7.29&160.24 &28.76&314.05 &39.59&491.83\\
& NodeNorm &6.64&43.90 &24.37&87.15 &40.00&136.59 &6.81&163.50 &25.29&363.83 &39.92&592.77 \\
& MeanNorm &4.99&42.56 &18.28&67.16 &29.72&95.27 &4.78&156.99 &19.04&266.17 &27.62&390.94 \\
& GroupNorm &13.76&62.60 &43.64&187.56 &53.86&336.75 &14.80&260.03 &41.11&848.65 &69.82&1550.90\\
& CombNorm &15.04&63.94 &41.08&207.56 &57.30&378.08 &15.76&266.54 &40.26&946.32 &78.38&1752.74 \\
\toprule
\multirow{1}{*}{Backbone} & \multirow{1}{*}{Settings} & \multicolumn{6}{c}{PubMed} & \multicolumn{6}{c}{OGBN-ArXiv} \\ \midrule
\multirow{6}{*}{GCN} & None &8.36&535.83 &64.80&1298.71 &130.28&2350.06 &133.14&5856.61 &1136.12&14058.69 &2295.50&23432.50 \\
& BatchNorm &8.49&535.84 &70.19&1586.98 &141.15&2945.82 &135.88&6022.00 &1178.30&16539.43 &2378.49&28559.35 \\
& PairNorm &10.08&535.83 &87.70&1589.46 &176.94&2951.15 &146.77&6021.99 &1317.47&16539.93 &2657.19&28560.36 \\
& NodeNorm &9.80&535.83 &85.75&1876.81 &173.00&3544.90 &146.87&6188.01 &1303.23&19029.60 &2631.94&33706.34 \\
& MeanNorm &8.70&535.83 &72.47&1298.71 &145.90&2350.88 &137.85&5856.61 &1191.41&14059.32 &2407.81&23433.75 \\
& GroupNorm &14.55&535.88 &141.71&3043.47 &286.61&5955.95 &231.48&7689.02 &2388.49&41540.97 & 4801.72&84847.76\\
& CombNorm &16.92&535.88 &162.72&3623.99 &330.01&7156.31 &245.18&8020.42 & 2444.32&43978.73 & 5204.64&95168.28\\
\midrule
\multirow{6}{*}{SGC} & None &8.99&424.60 &49.02&1066.76 &95.34&1800.10 &135.20&6232.20 &1004.12&12109.80 &2014.96&18826.87 \\
& BatchNorm &9.40&443.88 &54.32&1358.60 &105.91&2400.34 &140.39&6397.60 &1045.01&14589.29 &2085.69&23951.86 \\
& PairNorm &11.72&443.86 &72.73&1358.47 &142.71&2400.09 &158.84&6397.58 &1188.22&14589.16 &2377.52&23951.60 \\
& NodeNorm &11.49&463.19 &70.80&1645.54 &138.86&2996.44 &157.50&6563.60 &1181.12&17080.08 &2354.95&29098.83 \\
& MeanNorm &9.97&424.60 &56.75&1066.76 &110.74&1800.10 &142.81&6232.20 &1061.17&12109.80 &2127.01&18826.87 \\
& GroupNorm &18.86&578.84 &129.27&2814.14 &255.70&5412.07 &299.06&8064.47 &2321.44&39592.63 & 2334.56&37129.89\\
& CombNorm &21.56&618.91 &150.92&3387.71 &299.14&6597.04 &321.34&8395.86 & 1270.12&21882.53 & 2512.72&42276.51\\
\bottomrule
\end{tabular}}
\vspace{-2mm}
\end{table*}

\begin{table*}[!ht]
\caption{\small Training time per epoch (ms) and the max allocated memory (MB) for dropout tricks.}
\vspace{-4mm}
\label{tab:complex_dropout}
\centering
\resizebox{1\textwidth}{!}{
\begin{tabular}{@{}llcccccccccccc@{}}
\toprule
\multirow{3}{*}{Backbone} & \multirow{3}{*}{Settings} & \multicolumn{6}{c}{Cora} & \multicolumn{6}{c}{Citeseer} \\
\cmidrule(r){3-8} \cmidrule(r){9-14} 
& & \multicolumn{2}{c}{2} & \multicolumn{2}{c}{16} & \multicolumn{2}{c}{32} & \multicolumn{2}{c}{2} & \multicolumn{2}{c}{16} & \multicolumn{2}{c}{32} \\
\cmidrule(r){3-4} \cmidrule(r){5-6} \cmidrule(r){7-8} \cmidrule(r){9-10} \cmidrule(r){11-12} \cmidrule(r){13-14}
& & Time & Memory & Time & Memory & Time & Memory & Time & Memory & Time & Memory & Time & Memory \\
\midrule
\multirow{5}{*}{GCN} & None &3.09&191.92 &20.12&196.35 &30.36&201.40 &5.49&511.80 &20.54&529.13 &33.99&548.94 \\
& DropEdge &3.79&170.99 &20.24&174.85 &40.34&179.28 &6.09&461.56 &22.62&478.41 &40.66&505.25 \\
& DropNode &3.43&154.91 &18.91&158.51 &34.31&162.36 &5.71&427.82 &22.91&448.28 &39.12&498.90 \\
& FastGCN &3.68&168.32 &19.53&172.38 &36.08&176.61 &6.15&463.89 &22.79&480.76 &40.29&504.10 \\
& LADIES &5.60&168.54 &29.28&172.60 &55.82&179.19 &7.24&463.91 &34.28&480.76 &61.26&504.09 \\
\midrule
\multirow{5}{*}{SGC} & None &4.23&42.56 &10.95&67.16 &19.24&95.27 &3.96&156.99 &10.99&265.31 &18.12&390.08  \\
& DropEdge &4.36&41.65 &13.97&65.69 &26.89&93.18 &4.78&151.84 &15.07&263.68 &28.51&390.11 \\
& DropNode &3.30&40.94 &14.77&64.53 &25.77&91.64 &4.56&149.14 &17.08&257.70 &29.29&381.35  \\
& FastGCN &3.63&41.60 &14.67&65.61 &26.71&93.12 &4.71&151.82 &17.61&263.56 &31.78&390.04 \\
& LADIES &5.73&41.61 &26.12&65.61 &48.57&93.07 &5.68&151.80 &28.34&264.21 &52.29&390.40\\
\toprule
\multirow{1}{*}{Backbone} & \multirow{1}{*}{Settings} & \multicolumn{6}{c}{PubMed} & \multicolumn{6}{c}{OGBN-ArXiv} \\ \midrule
\multirow{5}{*}{GCN} & None &8.36&535.83 &64.80&1298.71 &130.28&2350.06 &133.14&5856.61 &1136.12&14058.69 &2295.50&23432.50 \\
& DropEdge &8.71&471.35 &64.95&1264.26 &130.22&2312.64 &111.98&4963.70 &977.04&13044.54 &1977.29&22277.59 \\
& DropNode &8.14&418.39 &61.52&1231.69 &120.77&2270.86 &102.11&4267.52 &842.81&12272.71 &1695.38&21375.29 \\
& FastGCN &9.52&497.76 &68.60&1277.93 &136.83&2326.76 &136.80&5590.19 &1108.86&13754.13 &2223.08&23082.86 \\
& LADIES &11.59&497.37 &95.30&1278.56 &192.06&2326.93 &158.54&5591.76 &1310.97&13753.46 &2622.87&23083.57 \\
\midrule
\multirow{5}{*}{SGC} & None &8.99&424.60 &49.02&1066.76 &95.34&1800.10 &135.20&6232.20 &1004.12&12109.80 &2014.96&18826.87 \\
& DropEdge &9.04&391.39 &49.26&1028.71 &94.95&1757.71 &116.09&5340.74 &850.09&11095.42 &1693.23&17672.46 \\
& DropNode &9.03&365.86 &45.43&1002.34 &86.39&1731.22 &106.58&4672.89 &713.64&10275.51 &1416.19&16794.15 \\
& FastGCN &10.09&407.25 &53.28&1043.69 &102.09&1772.71 &142.55&5966.34 &976.45&11803.59 &1940.84&18477.01 \\
& LADIES &12.32&407.25 &78.93&1043.35 &155.52&1772.43 &157.47&5966.72 &1176.25&11800.70 &2345.02&18474.28 \\
\bottomrule
\end{tabular}}
\vspace{-3mm}
\end{table*}

\vspace{-1em}
\section{Best Combo of Individual Tricks}
\label{sec: combining_tricks}
In the last section, we evaluate the different training tricks of each category fairly with ceteris paribus. However, being augmented with one specific trick, it is observed that deep GNNs often perform worse than their shallow variants  since they are \textbf{NOT trained well} to fully tackle the issues of over-smoothing, overfitting and gradient vanishing. Considering the standard training of other deep neural networks (e.g., CNNs~\cite{he2016deep} mixed with skip connection, normalization, dropout, learning rate decay, etc.), it is unpractical to stack deep GNNs with a standalone trick to get the desired performance. Based on the benchmark analysis, we are thus motivated to summarize the powerful trick combos and shred novel insight into the design philosophy of deep graph neural networks. 

\vspace{2mm}
\noindent
\begin{elaboration}
\textbf{The best trick combos}: Provided that the ``sweet point" hyperparameter configurations in Table~\ref{tab:experiment_settings} are adopted: \ding{182} On Cora, $64$ layers SGC with \textit{Initial Connection}, \textit{Dropout}, and \textit{Label Smoothing}; \ding{183} On PubMed, $32$ layers SGC with \textit{Jumping Connection} and \textit{No Dropout}; \ding{184} On Citeseer, $32$ layers GCN with \textit{Initial Connections}, \textit{Identity Mapping}, and \textit{Label Smoothing}; \ding{185} On OGBN-ArXiv, $16$ layer GCNII with \textit{Initial Connection}, \textit{Identity Mapping}, \textit{GroupNorm}, \textit{NodeNorm}, and \textit{Dropout}.
\end{elaboration}

\vspace{-1em}
\subsection{Our best combos of tricks and ablation study} 
We summarize our best trick combos for Cora, Citeseer, PubMed, and OGBN-ArXiv in the above box. Meanwhile, comprehensive ablation studies are conducted to support the superiority of our trick combos in Figure~\ref{fig:abaltion}. For each dataset, we examine $7\sim8$ combination variants by adding, removing or replacing certain tricks from our best trick combo. Results extensively endorse our carefully selected combinations.

\vspace{-1em}
\subsection{Comparison to state-of-the-art frameworks.} 
To further validate the effectiveness of our explored best combos, we perform comparisons with other previous state-of-the-art frameworks, including SGC~\cite{wu2019simplifying}, DAGNN~\cite{liu2020towards}, GCNII~\cite{chen2020simple}, JKNet~\cite{xu2018representation}, APPNP~\cite{klicpera2018predict}, GPRGNN~\cite{chien2021adaptive}. As shown in Table~\ref{tab:model_comparison}, we find that organically combining the existing training tricks consistently outperforms the previous elaborately crafted deep GCNs frameworks, on both small- and large-scale graph datasets. 

\vspace{-1em}
\subsection{Transferring trick combos across datasets.} 

The last sanity check is whether there exist certain trick combos can be effective across multiple different graph datasets. Therefore, we pick \textit{Initial Connection} + \textit{GroupNorm} as the trick combo since these two techniques achieve optimal performance under most of scenarios in Section~\ref{sec:methods_connection},~\ref{sec:methods_norm}, and~\ref{sec:methods_dropping}, and this trick combo also performs on-par with the best trick combo on Cora and OGBN-ArXiv. Specially, we evaluate it on \textbf{eight} other open-source graph datasets: ($i$) two Co-author datasets~\cite{shchur2018pitfalls} (CS and Physics), ($ii$) two Amazon datasets~\cite{shchur2018pitfalls} (Computers and Photo), ($iii$) three WebKB datasets~\cite{pei2020geom} (Texas, Wisconsin, Cornell), and ($iv$) the Actor dataset~\cite{pei2020geom}. In these transfer investigations, we follow the exact ``sweet point'' settings from Cora in Table~\ref{tab:experiment_settings}, except that for two Coauthor datasets, the weight decay is set to $0$ and the dropout rate is set to $0.8$, and in two Amazon datasets the dropout rate is set to $0.5$, as adopted by~\cite{liu2020towards}.


As shown in Table~$8$, ($i$) our chosen trick combo (i.e., \textit{Initial Connection} + \textit{GroupNorm}) \textbf{\textit{universally}} \textbf{encourages the worst-performing SGC to become one of the top-performing candidates}. It consistently surpasses SGC, DAGNN, GCNII, JKNet, APPNP, GPRGNN, and DGMLP by a substantial accuracy margin up to $61.69\%$ improvements, except for the comparable performances on CS, Computers, and Photo graph datasets. The superior transfer performance of our trick combo suggests that it is capable of serving as a stronger baseline for future deep GNN researches. ($ii$) Meanwhile, we observe that equipping our proposed training tricks improves the stability of SGC (i.e., fewer performance variances) on some graph datasets like CS, Photo, Texas, Wisconsin, and Actor, while incurring degraded stability on the rest of the three datasets. ($iii$) Our trick combo transfers better on small-scale graph datasets such as Photo, Texas, Wisconsin, Cornell, and Actor, compared to other three larger graph datasets like CS, Physics, and Computers.

\vspace{-0.6em}
\section{Extra Analysis}
\vspace{-0.4em}
\subsection{Hyperparameter Analysis}
We consider the key hyperparameters of \{learning rate, weight decay, dropout rate, and hidden dimension\}, and study their impacts on the deep GNN training. Specifically, a 16-layer GCNII is adopted as the backbone architecture since it normally achieves promising performance in deep GNNs literature. As shown in Figure~\ref{fig:hp}, from the results on Cora, Citeseer, Pubmed, and OGBN-ArXiv, we find that: ($i$) The optimal dropout rate varies across different scales of datasets; ($ii$) The weight decay for hidden layers plays a crucial role in deep GNNs, especially in the over-parameterized context (e.g., deep GNNs on small-scale dataset like Cora). ($iii$) It seems that the deep GNNs' performance is not sensitive to the hidden dimension, and changing it from $128$ to $512$ only has moderated effects. ($iv$) In general, our identified ``sweet point" of hyperparameters achieves superior results, which demonstrates its rationality as a common configuration to benchmark the deep GNN training tricks.

\vspace{-1em}
\subsection{Complexity Analysis}
To analyze the time and space complexity in practice, we conduct experiments on 2/16/32 layers for GNNs with different tricks. From the results on Table~\ref{tab:complex_skip}, ~\ref{tab:complex_norm} and~\ref{tab:complex_dropout}, we summarize our observations for different types of tricks as follows. All results are measured on a single Quadro RTX 8000, with \texttt{Linux} + \texttt{cuda driver 11.2} + \texttt{PyTorch/torch geometric/torch sparse/etc with cuda 11.1}.

\textbf{Skip connections.} As shown in Table~\ref{tab:complex_skip}, we can draw the following observations: $(i)$ Residual, initial and jumping connections need negligibly additional training time and on-par memory space compared with vanilla GNNs. It is attributed to that only some constant operations and valid extra memory space are required in skip connections, e.g. initial connection only requires to store the embeddings from the first linear layer and perform several times of adding operation in addition. $(ii)$ Dense connection requires approximately twice the training time and memory space of vanilla GCN, which is apparently impractical when dataset and model go larger.

\textbf{Normalization.} As shown in Table~\ref{tab:complex_norm}, the observations on normalization tricks are: $(i)$ Graph Neural Networks with activation-based normalization layers (i.e. BatchNorm, NodeNorm and MeanNorm) have on-par time and space complexity with vanilla GNNs. $(ii)$ Parametric normalization layers (i.e. PairNorm, GroupNorm and CombNorm) require more training time and memory space. It is because these normalization layers have more learnable parameters and requires a number of extra inference time to distinguish node embeddings and mitigate the over-smoothing problem.

\textbf{Dropout.} As shown in Table~\ref{tab:complex_dropout}, we have the following findings: $(i)$ To prune the graph structure with specified strategies, dropout tricks requires a valid period of time for pruning. Such a pruning procedure is performed every training epoch, and thus time consuming. $(ii)$ With sparsified graph structure, dropout tricks moderately reduce the memory consumption. 

\vspace{-1em}
\section{Conclusion} \label{sec:conclusion}
Deep graph neural networks (GNNs) is a promising field that has so far been a bit held back by finding out ``truly" effective training ticks to alleviate the notorious over-smoothing issue. This work provides a standardized benchmark with fair and consistent experimental configurations to push this field forward. We broadly investigate dozens of existing approaches on tens of representative graphs. Based on extensive results, we identify the combo of most powerful training tricks, and establish new sate-of-the-art performance for deep GNNs. We hope them to lay a solid, fair, and practical evaluation foundations by providing strong baselines and superior trick combos for deep GNN research community.

\vspace{-1em}
\section*{Acknowledgements}
X. Hu is in part supported by NSF (\#IIS-1750074, \#IIS-1900990, \#IIS-1849085). Z. Wang is in part supported by US Army Research Office Young Investigator Award W911NF2010240.

\vspace{-1em}


{\small
  \bibliographystyle{IEEEtran}
  \bibliography{conferences_abrv,GCNB}
}


\vspace{-15mm}
\begin{IEEEbiography}[{\includegraphics[width=1in,height=1.25in,clip,keepaspectratio]{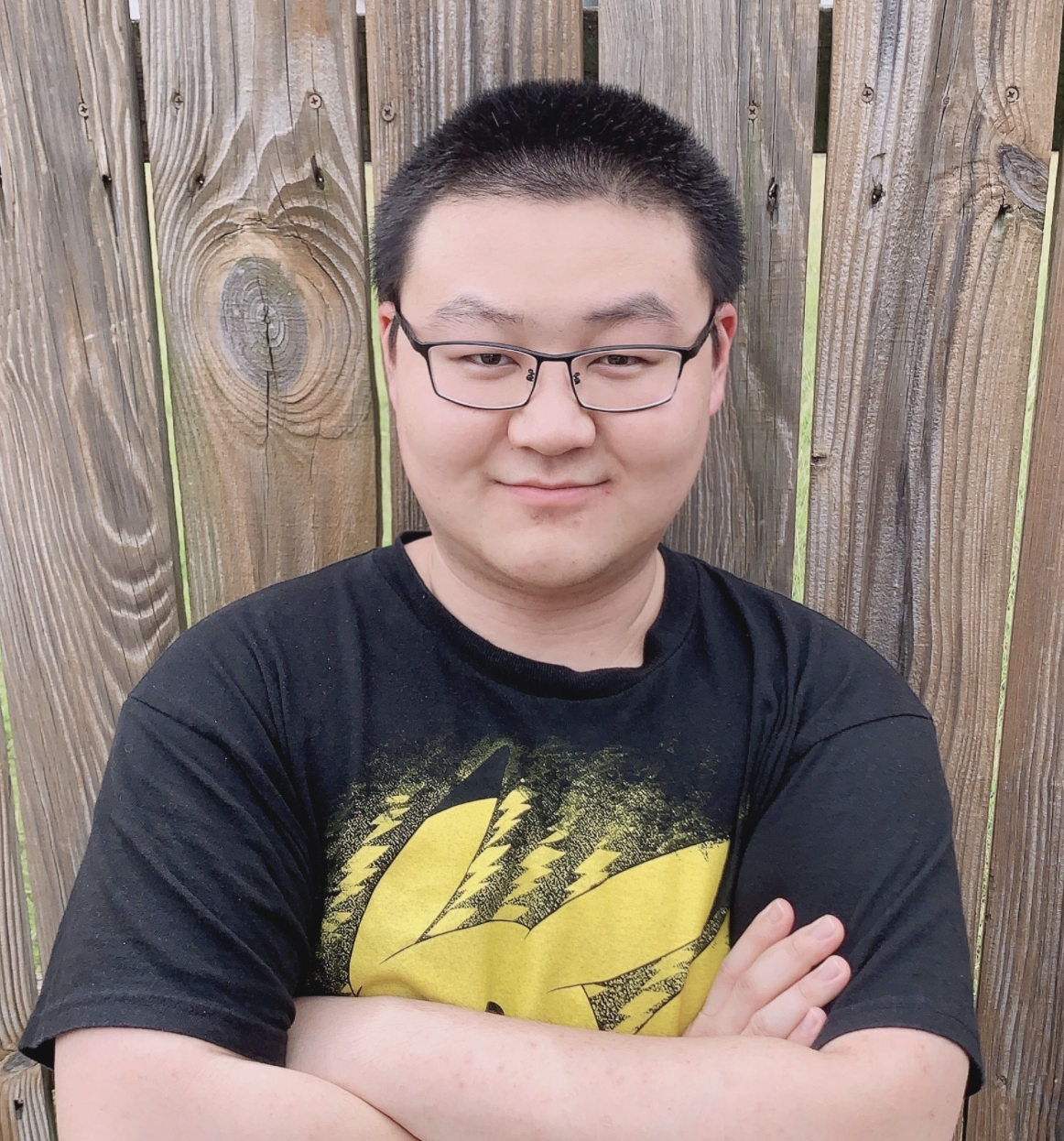}}]{Tianlong Chen} is a Ph.D. student in Electrical and Computer Engineering at University of Texas at Austin. He received his B.S. degree from University of Science and Technology of China in 2017. He has also interned at IBM Research, Facebook AI, Microsoft Research and Walmart Technology Lab. His research focuses sparse neural networks, AutoML, adversarial robustness, self-supervised learning and graph learning.
\end{IEEEbiography}
\vspace{-15mm}

\begin{IEEEbiography}[{\includegraphics[width=1in,height=1.1in,clip,keepaspectratio]{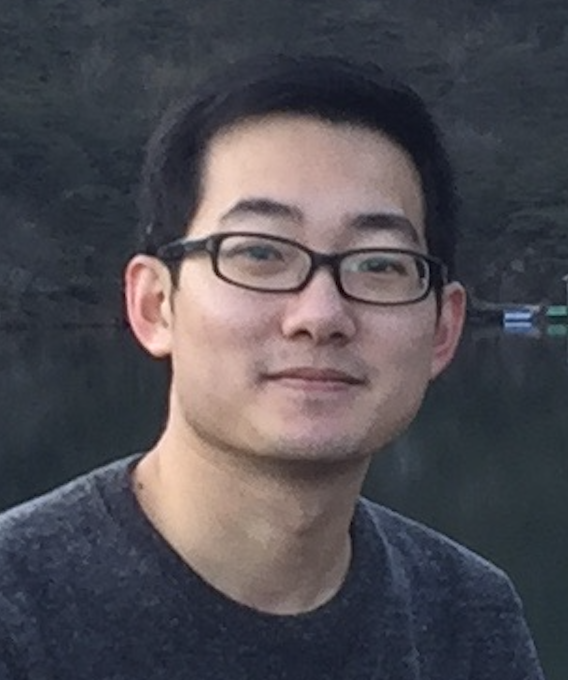}}]{Kaixiong Zhou} received the BS and MS degrees in electrical engineering and information science from Sun Yat-Sen University and University of Science and Technology of China, respectively. He is currently a Ph.D student in the Department of Computer Science at Rice University. His research interests include data mining, network analytics, and graph neural networks. 
\end{IEEEbiography}
\vspace{-15mm}

\begin{IEEEbiography}[{\includegraphics[width=1.1in,height=1.1in,clip,keepaspectratio]{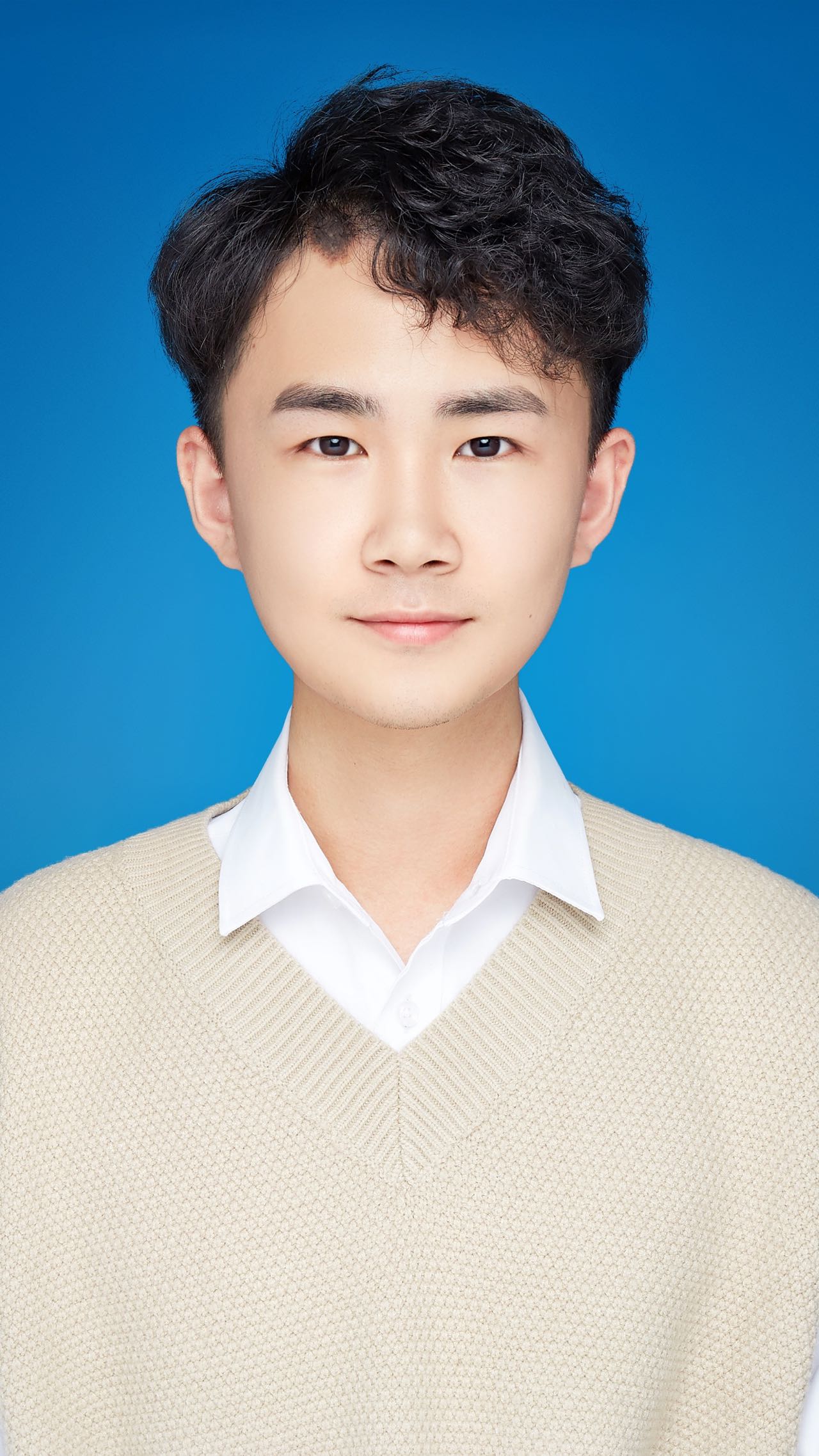}}]{Keyu Duan} is currently pursuing his Ph.D. degree in computer science with Rice Univeristy, supervised by Dr. Xia (Ben) Hu at DATA lab. Before that, he received the B.E. degree of Software Engineering from Beihang University, Beijing, China. His research interests include Network Analytics, Graph Neural Networks, Scalable Graph Representation Learning, Knowledge Graph Reasoning.
\end{IEEEbiography}
\vspace{-15mm}

\begin{IEEEbiography}[{\includegraphics[width=1.25in,height=1.25in,clip,keepaspectratio]{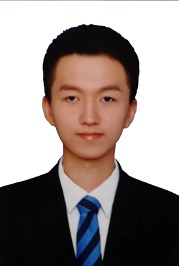}}]{Wenqing Zheng} received his B.Eng. degree in Communication Engineering from Beijing University of Posts and Telecommunications, China, in 2018, he received his M.S. in Eng. At the University of Texas at Austin in Dec. 2020. He is currently a Ph.D. student in the Department of Electrical and Computer Engineering at the University of Texas at Austin. His research interests include graph neural networks, transformers and and symbolic reasoning.
\end{IEEEbiography}
\vspace{-15mm}

\begin{IEEEbiography}[{\includegraphics[width=1.0in,height=1.25in,clip,keepaspectratio]{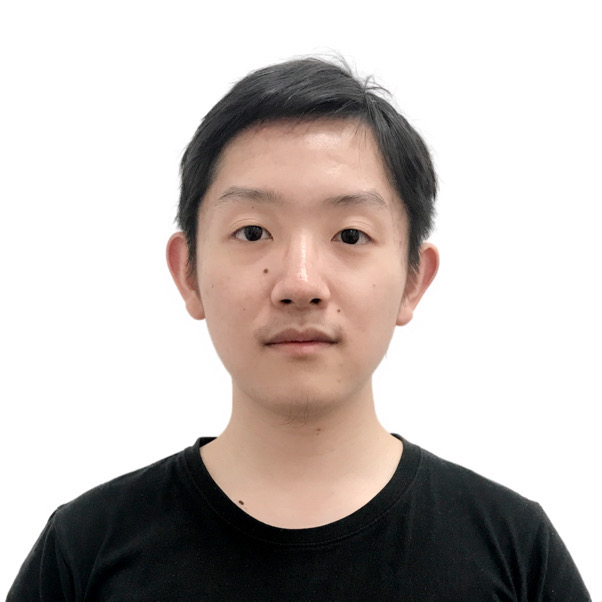}}]{Peihao Wang} is currently pursuing his Ph.D. degree in Electrical and Computer Engineering at the University of Texas at Austin. Prior to that, he received the B.E. degree of Computer Science from ShanghaiTech University, China, in 2020. His research interests focus on graph representation learning, 3D vision and graphics, and computational photography.
\end{IEEEbiography}
\vspace{-15mm}

\vspace{-3mm}
\begin{IEEEbiography}[{\includegraphics[width=1.in,height=1.2in,clip,keepaspectratio]{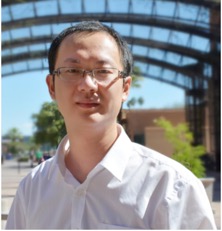}}]{Xia Hu} received the BS and MS degrees in computer science from Beihang University, China and the PhD degree in computer science and engineering from Arizona State University. He is currently an Associate Professor with the Department of Computer Science, Rice University. He has published nearly $100$ papers in several major academic venues, including NeurIPS, ICLR, ICML, etc. His developed automated machine learning package, AutoKeras, has become the most rated open-source AutoML system.
\end{IEEEbiography}
\vspace{-12mm}

\begin{IEEEbiography}[{\includegraphics[trim={0 10cm 0 3cm},width=1in,height=1.25in,clip,keepaspectratio]{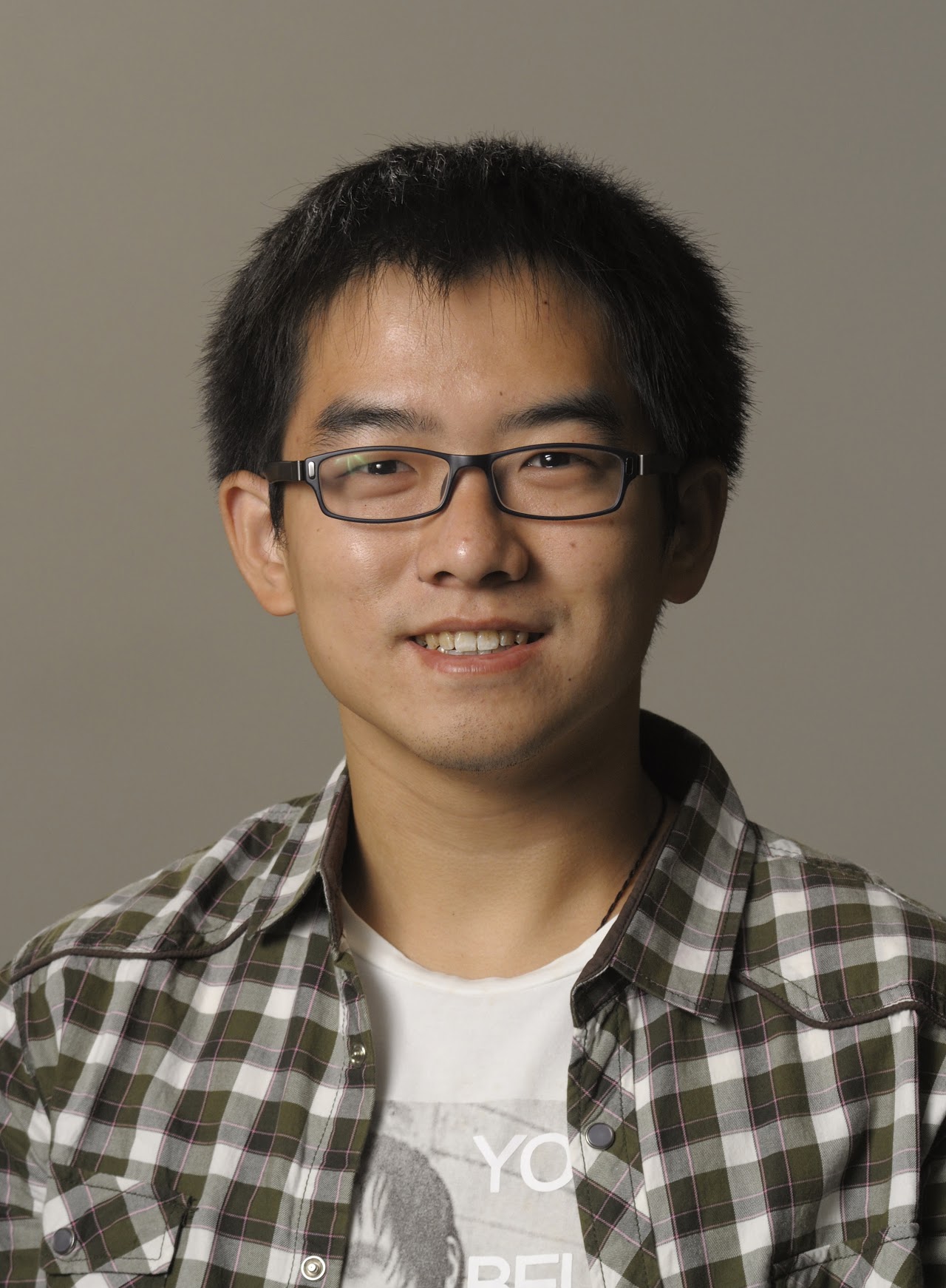}}]{Zhangyang Wang}
is currently an Assistant Professor of ECE at UT Austin. He received his Ph.D. in ECE from UIUC in 2016, and his B.E. in EEIS from USTC in 2012. Prof. Wang is broadly interested in the fields of machine learning, computer vision, optimization, and their interdisciplinary applications. His latest interests focus on automated machine learning (AutoML), learning-based optimization, machine learning robustness, and efficient deep learning.
\end{IEEEbiography}

\clearpage

\appendices

\section{More Technical Details} \label{sec:more_tech_details}

\textbf{Diverse hyperparameter configurations of existing deep GNNs.} As shown in Table~\ref{tab:settings_cite},~\ref{tab:settings_pubmed}, and~\ref{tab:settings_arxiv}, we can see that these highly inconsistent hyperparameter settings pose severely challenges to fairly compare the existing training tricks for deep GNNs. 

\vspace{-4mm}
\begin{table}[!ht]
\caption{\small Configurations adopted on Citeseer~\cite{kipf2016semi}.}
\vspace{-4mm}
\label{tab:settings_cite}
\centering
\resizebox{0.40\textwidth}{!}{
\begin{tabular}{@{}lrrrrr@{}}
\toprule
Methods & Total epoch & Learning rate \& Decay & Weight decay & Dropout & Hidden dimension \\
\midrule
Chen et al. (2020)~\cite{chen2020simple} & 100 & 0.01 & 5e-4 & 0.6 & 256 \\
Xu et al. (2018)~\cite{xu2018representation} & - & 0.005 & 5e-4 & 0.5 & \{16, 32\} \\
Klicpera et al. (2018)~\cite{klicpera2018predict} & 10000 & 0.01 & 5e-3 & 0.5 & 64  \\
Zhang et al. (2020)~\cite{zhang2020revisiting} & 1500 & \{0.001, 0.005, 0.01\} & - & \{0.1, 0.2, 0.3, 0.4, 0.5\} & 64 \\
Luan et al. (2019)~\cite{luan2019break} & 3000 & 2.8218e-03 & 1.9812e-02 & 0.98327 & 5000 \\
Liu et al. (2020)~\cite{liu2020towards} & 100 & 0.01 & 0.005 & 0.8 & 64 \\
Zhao et al. (2019)~\cite{zhao2019pairnorm} & 1500 & 0.005 & 5e-4 & 0.6 & 32 \\
Min et al. (2020)~\cite{scattering1} & 200 & 0.005 & 0 & 0.9 & - \\
Zhou et al. (2020)~\cite{zhou2020towards} & 1000 & 0.005 & 5e-4 & 0.6 & - \\
Zhou et al. (2020)~\cite{zhou2020understanding} & 50 & 0.005 & 1e-5 & 0 & - \\
Rong et al. (2020)~\cite{rong2020dropedge} & 400 & 0.009 & 1e-3 & 0.8 & 128 \\
Zou et al. (2020)~\cite{zou2019layer} & 100 & 0.001 & - & - & 256 \\
Hasanzadeh et al. (2020)~\cite{hasanzadeh2020bayesian} & 1700 & 0.005 & 5e-3 & 0 & 128 \\
\bottomrule
\end{tabular}}
\vspace{-1mm}
\end{table}

\vspace{-6mm}
\begin{table}[!ht]
\caption{\small Configurations adopted on PubMed~\cite{kipf2016semi}.}
\vspace{-4mm}
\label{tab:settings_pubmed}
\centering
\resizebox{0.40\textwidth}{!}{
\begin{tabular}{@{}lrrrrr@{}}
\toprule
Methods & Total epoch & Learning rate \& Decay & Weight decay & Dropout & Hidden dimension \\
\midrule
Chen et al. (2020)~\cite{chen2020simple} & 100 & 0.01 & 5e-4 & 0.5 & 256 \\
Klicpera et al. (2018)~\cite{klicpera2018predict} & 10000 & 0.01 & 5e-3 & 0.5 & 64 \\
Zhang et al. (2020)~\cite{zhang2020revisiting} & 1500 & \{0.001, 0.005, 0.01\} & - & \{0.1, 0.2, 0.3, 0.4, 0.5\} & 64 \\
Luan et al. (2019)~\cite{luan2019break} & 3000 & 0.001 & 0.02 & 0.65277 & 128 \\
Liu et al. (2020)~\cite{liu2020towards} & 100 & 0.01 & 0.005 & 0.8 & 64 \\
Zhao et al. (2019)~\cite{zhao2019pairnorm} & 1500 & 0.005 & 5e-4 & 0.6 & \{32, 64\} \\
Zhou et al. (2020)~\cite{zhou2020towards} & 1000 & 0.01 & 1e-3 & 0.6 & - \\
Rong et al. (2020)~\cite{rong2020dropedge} & - & 0.01 & 1e-3 & 0.5 & 128 \\
Zou et al. (2020)~\cite{zou2019layer} & 100 & 0.001 & - & - & 256 \\
Hasanzadeh et al. (2020)~\cite{hasanzadeh2020bayesian} & 2000 & 0.005 & 5e-3 & 0 & 128 \\
\bottomrule
\end{tabular}}
\vspace{-1mm}
\end{table}

\vspace{-6mm}
\begin{table}[!ht]
\caption{\small Configurations adopted on Ogbn-ArXiv~\cite{hu2020open}.}
\vspace{-4mm}
\label{tab:settings_arxiv}
\centering
\resizebox{0.40\textwidth}{!}{
\begin{tabular}{@{}lrrrrr@{}}
\toprule
Methods & Total epoch & Learning rate \& Decay & Weight decay & Dropout & Hidden dimension\\
\midrule
Xu et al. (2018)~\cite{xu2018representation} & 500 & 0.01 & - & 0.5 & 128 \\
Li et al. (2019, 2020)~\cite{li2019deepgcns,li2020deepergcn} & 500 & 0.01 & - & 0.5 & 128 \\
Chen et al. (2020)~\cite{chen2020simple} & 1000 & 0.001 & - & 0.1 & 256 \\
Zhang et al. (2020)~\cite{zhang2020revisiting} & 1500 & \{0.001, 0.005, 0.01\} & - & {0.1, 0.2, 0.3, 0.4, 0.5} & 256 \\
Liu et al. (2020)~\cite{liu2020towards} & 1000 & 0.005 & - & 0.2 & 256 \\
Zhou et al. (2020)~\cite{zhou2020understanding} & 400 & 0.005 & 0.001 & 0.6 & - \\
Shi et al. (2020)~\cite{shi2020masked} & 2000 & 0.001 & 0.0005 & 0.3 & 128 \\
Kong et al. (2020)~\cite{kong2020flag} & 500 & 0.01 & - & 0.5 & 256 \\
Sun et al. (2020)~\cite{sun2020acne} & 2000 & 0.002 & - & 0.75 & 256 \\
\bottomrule
\end{tabular}}
\vspace{-1mm}
\end{table}

\vspace{-6mm}
\section{More Implementation Details} \label{sec:more_imp_details}

\textbf{Datasets and computing facilities.} Table~\ref{table:datasets} provides the detailed properties and download links for all adopted datasets. We adopt the following benchmark datasets since i) they are widely applied to develop and evaluate GNN models, especially for deep GNNs studied in this paper; ii) they contains diverse graphs from small-scale to large scale or from homogeneous to heterogeneous; iii) they are collected from different applications including citation network, social network, etc. Specifically, the four datasets applied in benchmark studies are not the ``small-world'' graphs where any node can reach the other nodes in a few hops. The maximum diameters of strongly connected components in Cora, Citeseer, Pubmed, OGBN-arxiv are 19, 28, 18, and 23, respectively. The deep GNNs help encode the high-order neighborhood dependencies and the complex neighbor substructure for each node, which could boost the node classification tasks. All experiments on large graph datasets, e.g., OGBN-ArXiv, are conducted on single 48G Quadro RTX 8000 GPU. For other experiments on small graphs such as Cora, are ran at singe 12G GTX 2080TI GPU. The detailed descriptions of these datasets are listed in the following. 
\begin{itemize}
    \item \textbf{Cora, Citeseer, Pubmed.} They are the scientific citation network datasets~\cite{yang2016revisiting, kipf2016semi}, where nodes and edges represent the scientific publications and  their citation relationships, respectively. Each node is described by bag-of-words representation, i.e., the 0/1-valued word vector indicating the absence/presence of the corresponding words in the scientific publication. Each node is associated with a one-hot label, where the node classification task is to predict what field the corresponding publication belongs to. 
    \item \textbf{OGBN-Arxiv.} The OGBN-Arxiv dataset is a benchmark citation network collected in open graph benchmark (OGB)~\cite{hu2020open}, which has been widely to evaluate GNN models recently. Each node represents an arXiv paper from computer science domain, and each directed edge indicates that one paper cites another one. The node is described by a 128-dimensional word embedding extracted from the title and abstract in the corresponding publication. Similar to  Cora, Citeseer, and Pubmed, the node classification task is to predict the subject areas of the corresponding arXiv papers. 
    \item \textbf{Coauthor CS, Coauthor Physics.} They are the co-authorship graph datasets~\cite{shchur2018pitfalls} from the scientific fields of computer science and physics, respectively. The nodes represent the authors, and the links indicate whether the two corresponding nodes co-authored papers. Node features represent paper keywords for each author’s papers. The node classification task  is to predict the most active fields of study for the corresponding author.
    \item \textbf{Amazon Computers, Amazon Photo.} These two datasets~\cite{shchur2018pitfalls} are the segments of the Amazon co-purchase graph~\cite{mcauley2015image}. While nodes are the items, edges indicate whether the two items re frequently bought together or not. The node features are given by bag-of-words representations extracted from the product reviews, and the class labels denote the product categories.
    \item \textbf{Texas, Wisconsin, Cornell.} They are the three subdatasets of WebKB (\url{http://www.cs.cmu.edu/afs/cs.cmu.edu/project/theo-11/www/wwkb}) collected from computer science departments of various universities by Carnegie Mellon University~\cite{pei2020geom}. While nodes represent webpages in the webpage datasets, edges are hyperlinks between them. The node feature vectors are given by bag-of-word representation of the corresponding webpages. Each node is associated with one-hot label to indicate one of the following five categories, i.e., student, project, course, staff, and faculty.
    \item \textbf{Actor.} This is an actor co-occurrence network, which is an actor-only induced subgraph of the film-director-actor-writer network~\cite{tang2009social}. In the actor co-occurrence network, nodes correspond to actors and edges denote the co-occurrence relationships on the same Wikipedia pages. Node feature vectors are described by bag-of-word representation of keywords in the actors’ Wikipedia pages. The node classification task is to predict the topic of the actor’s Wikipedia page.
\end{itemize}

\textbf{Backbone Details.} We implement backbones GCN (\url{https://github.com/tkipf/gcn}) and SGC (\url{https://github.com/Tiiiger/SGC}) to compare the different tricks as studied in Section~\ref{sec:tricks_study}. Specifically, we follow the official model implementations of GCN and SGC to evaluate tricks of graph normalization and random dropping. While the default dropout rate listed in Table~\ref{tab:experiment_settings} is applied for backbones (GCN/SGC+NoNorm) in benchmarking normalization techniques, the dropout rate is optimized from set $\{0.2, 0.5, 0.7\}$ for backbones (GCN/SGC+Dropout) in benchmarking dropping methods. To evaluate the skip connections and identity mapping, we further include feature transformation layers to the initial and the end of backbones, which project initial features into hidden embedding space and map hidden embeddings to node labels, respectively. The backbones are denoted by GCN/SGC+None and GCN+without in Tables~\ref{tab:res_connection} and \ref{tab:other_trick}, respectively. The slight adaptions over the backbones explain the different baseline performances for the benchmark studies of skip connections, graph normalization, random dropping, and others. 

\begin{table}[htb]
\centering
\vspace{-4mm}
\caption{Graph datasets statistics and download links.}
\vspace{-4mm}
\label{table:datasets}
\resizebox{0.41\textwidth}{!}{
\begin{tabular}{@{}lrrrrrl@{}}
\toprule
Dataset & Nodes & Edges & Ave. Degree & Features & Classes & Download Links\\ 
\midrule
Cora & 2,708 &  5,429 &  3.88 &  1,433  &  7 & \url{https://github.com/kimiyoung/planetoid/raw/master/data}\\ \midrule
Citeseer  &  3,327 &  4,732 &  2.84 &  3,703  &  6 & \url{https://github.com/kimiyoung/planetoid/raw/master/data} \\ \midrule
PubMed &  19,717 &  44,338 &  4.50 & 500  & 3 & \url{https://github.com/kimiyoung/planetoid/raw/master/data}\\ \midrule
OGBN-ArXiv & 169,343 & 1,166,243 & 13.77 & 128 & 40 & \url{https://ogb.stanford.edu/}\\ \midrule
Coauthor CS & 18,333 & 81,894 & 8.93 & 6805 & 15 & \url{https://github.com/shchur/gnn-benchmark/raw/master/data/npz/}\\ \midrule
Coauthor Physics &  34,493 & 247,962 & 14.38 & 8,415 & 5 & \url{https://github.com/shchur/gnn-benchmark/raw/master/data/npz/}\\ \midrule
Amazon Computers & 13,381 & 245,778 & 36.74 & 767 & 10 & \url{https://github.com/shchur/gnn-benchmark/raw/master/data/npz/}\\ \midrule
Amazon Photo & 7,487 & 119,043 & 31.80 & 745 & 8 & \url{https://github.com/shchur/gnn-benchmark/raw/master/data/npz/}\\ \midrule
Texas & 183 & 309 & 3.38 & 1,703 & 5 & \url{https://raw.githubusercontent.com/graphdml-uiuc-jlu/geom-gcn/master}\\ \midrule
Wisconsin & 183 & 499 & 5.45 & 1,703 & 5 & \url{https://raw.githubusercontent.com/graphdml-uiuc-jlu/geom-gcn/master}\\ \midrule
Cornell & 183 & 295 & 3.22 & 1,703 & 5 & \url{https://raw.githubusercontent.com/graphdml-uiuc-jlu/geom-gcn/master}\\ \midrule
Actor & 7,600 & 33,544 & 8.83 & 931 & 5 & \url{https://raw.githubusercontent.com/graphdml-uiuc-jlu/geom-gcn/master}\\
\bottomrule
\end{tabular}}
\vspace{-4mm}
\end{table}

\vspace{-3mm}
\section{More Experimental Results} \label{sec:more_exp_results}

\textbf{More results of skip connections.} Table~\ref{tab:res_connection_full} collects the achieved test accuracy under different skip connection mechanisms, and the standard deviations of $100$ independent runs are reported.

\vspace{-4mm}
\begin{table}[!ht]
\caption{Test accuracy (\%) with skip connection.}
\vspace{-4mm}
\label{tab:res_connection_full}
\centering
\resizebox{0.48\textwidth}{!}{
\begin{tabular}{@{}llcccccccccccc@{}}
\toprule
\multirow{2}{*}{Backbone} & \multirow{2}{*}{Settings} & \multicolumn{3}{c}{Cora} & \multicolumn{3}{c}{Citeseer} & \multicolumn{3}{c}{PubMed} & \multicolumn{3}{c}{OGBN-ArXiv}\\
\cmidrule(r){3-5} \cmidrule(r){6-8} \cmidrule(r){9-11} \cmidrule(r){12-14}
& & 2 & 16 & 32 & 2 & 16 & 32 & 2 & 16 & 32 & 2 & 16 & 32\\ 
\midrule
\multirow{5}{*}{GCN} & Residual & 74.73$\pm$3.41 & 20.05$\pm$4.66 & 19.57$\pm$4.94 & 66.83$\pm$1.37 & 20.77$\pm$1.25 & 20.90$\pm$1.34 & 75.27$\pm$1.72 & 38.84$\pm$1.39 & 38.74$\pm$1.33 & 70.19$\pm$0.17 & 69.34$\pm$0.30 & 65.09$\pm$1.08\\
& Initial & 79.00$\pm$1.01 & \textbf{78.61$\pm$0.84} & \textbf{78.74$\pm$0.86} & 70.15$\pm$0.84 & \textbf{68.41$\pm$0.93} & \textbf{68.36$\pm$1.10} & 77.92$\pm$1.76 & \textbf{77.52$\pm$1.55} & \textbf{78.18$\pm$0.49} & 70.16$\pm$0.21 & 70.50$\pm$0.28 & 70.23$\pm$0.26\\
& Jumping & 80.98$\pm$0.85 & 76.04$\pm$3.04 & 75.57$\pm$3.74 & 69.33$\pm$0.92 & 58.38$\pm$5.53 & 55.03$\pm$6.27 & 77.83$\pm$0.88 & 75.62$\pm$1.95 & 75.36$\pm$2.48 & \textbf{70.24$\pm$0.23} & \textbf{71.83$\pm$0.26} & \textbf{71.87$\pm$0.23}\\
& Dense   & 77.86$\pm$1.73 & 69.61$\pm$4.31 & 67.26$\pm$6.51 & 66.18$\pm$1.93 & 49.33$\pm$7.79 & 41.48$\pm$7.85 & 72.53$\pm$2.61 & 69.91$\pm$6.95 & 62.99$\pm$10.21 & 70.08$\pm$0.24 & 71.29$\pm$0.23 & 70.94$\pm$0.30\\
& None    & \textbf{82.38$\pm$0.33} & 21.49$\pm$3.84 & 21.22$\pm$3.71 & \textbf{71.46$\pm$0.44} & 19.59$\pm$1.96 & 20.29$\pm$1.79 & \textbf{79.76$\pm$0.39} & 39.14$\pm$1.38 & 38.77$\pm$1.20 & 69.46$\pm$0.22 & 67.96$\pm$0.38 & 45.48$\pm$4.50\\
\midrule
\multirow{5}{*}{SGC} & Residual & \textbf{81.77$\pm$0.28} & 82.55$\pm$0.41 & 80.14$\pm$0.40 & 71.68$\pm$0.33 & 71.31$\pm$0.57 & 71.00$\pm$0.49 & 78.87$\pm$0.29 & 79.86$\pm$0.25 & 79.07$\pm$0.35 & 69.09$\pm$0.13 & 66.52$\pm$0.23 & 61.83$\pm$0.36\\
& Initial & 81.40$\pm$0.26 & \textbf{83.66$\pm$0.38} & 83.77$\pm$0.38 & 71.60$\pm$0.33 & \textbf{72.16$\pm$0.30} & \textbf{72.25$\pm$0.38} & \textbf{79.11$\pm$0.23} & 79.73$\pm$0.23 & 79.74$\pm$0.24 & 68.93$\pm$0.11 & 69.24$\pm$0.16 & 69.15$\pm$0.17\\
& Jumping & 77.75$\pm$0.65 & 83.42$\pm$0.50 & \textbf{83.88$\pm$0.48} & 69.96$\pm$0.37 & 71.89$\pm$0.52 & 71.88$\pm$0.58 & 77.42$\pm$0.30 & \textbf{79.99$\pm$0.46} & \textbf{80.07$\pm$0.67} & 68.76$\pm$0.17 & 70.61$\pm$0.19 & 70.65$\pm$0.23\\
& Dense   & 77.31$\pm$0.39 & 81.24$\pm$1.12 & 77.66$\pm$2.74 & 70.99$\pm$0.58 & 67.75$\pm$1.85 & 66.35$\pm$5.61 & 77.12$\pm$0.73 & 72.77$\pm$5.12 & 74.84$\pm$1.58 & \textbf{69.39$\pm$0.18} & \textbf{71.42$\pm$0.28} & \textbf{71.52$\pm$0.31} \\
& None    & 79.31$\pm$0.37 & 75.98$\pm$1.06 & 68.45$\pm$3.10 & \textbf{72.31$\pm$0.38} & 71.03$\pm$1.18 & 61.92$\pm$3.48 & 78.06$\pm$0.31 & 69.18$\pm$0.58 & 66.61$\pm$0.56 & 61.98$\pm$0.08 & 41.58$\pm$0.27 & 34.22$\pm$0.04\\
\bottomrule
\end{tabular}}
\vspace{-2mm}
\end{table}

\textbf{More results of graph normalizations.} Table~\ref{tab:norm_full} reports the performance with different graph normalization techniques. The standard deviations are included.

\vspace{-4mm}
\begin{table}[!ht]
\caption{Test accuracy (\%) with graph normalization.}
\vspace{-4mm}
\label{tab:norm_full}
\centering
\resizebox{0.48\textwidth}{!}{
\begin{tabular}{@{}llcccccccccccc@{}}
\toprule
\multirow{2}{*}{Backbone} & \multirow{2}{*}{Settings} & \multicolumn{3}{c}{Cora} & \multicolumn{3}{c}{Citeseer} & \multicolumn{3}{c}{PubMed} & \multicolumn{3}{c}{OGBN-ArXiv}\\
\cmidrule(r){3-5} \cmidrule(r){6-8} \cmidrule(r){9-11} \cmidrule(r){12-14}
& & 2 & 16 & 32 & 2 & 16 & 32 & 2 & 16 & 32 & 2 & 16 & 32\\ 
\midrule
\multirow{5}{*}{GCN} 
 & BatchNorm & 69.91$\pm$1.96 & 61.20$\pm$13.88 & \textbf{ 29.05$\pm$10.68} & 46.27$\pm$2.30 & 26.25$\pm$12.11 & 21.82$\pm$7.91 & 67.15$\pm$1.94 & 58.00$\pm$13.42 & 55.98$\pm$12.98 & 70.44$\pm$0.22 & 70.52$\pm$0.46 & 68.74$\pm$0.84\\
& PairNorm & 74.43$\pm$1.36 & \textbf{55.75$\pm$13.19} & 17.67$\pm$9.23 & 63.26$\pm$1.32 & \textbf{27.45$\pm$7.22} & 20.67$\pm$6.51 & 75.67$\pm$0.86 & 71.30$\pm$3.00 & 61.54$\pm$13.80 & 65.74$\pm$0.16 & 65.37$\pm$0.69 & 63.32$\pm$0.97\\
 & NodeNorm & 79.87$\pm$0.80 & 21.46$\pm$9.50 & 21.48$\pm$9.41 & 68.96$\pm$1.00 & 18.81$\pm$2.12 & 19.03$\pm$2.35 & 78.14$\pm$0.79 & 40.92$\pm$0.29 & 40.93$\pm$0.29 & 70.62$\pm$0.23 & 70.75$\pm$0.43 & 29.94$\pm$2.29\\
 & MeanNorm & \textbf{82.49$\pm$0.35} & 13.51$\pm$3.14 & 13.03$\pm$0.15 & 70.86$\pm$0.36 & 16.09$\pm$3.40 & 7.70$\pm$0.00 & 78.68$\pm$0.41 & 18.92$\pm$3.20 & 18.00$\pm$0.00 & 69.54$\pm$0.15 & 70.40$\pm$0.25 & 56.94$\pm$20.26\\
 & GroupNorm & 82.41$\pm$0.33 & 41.76$\pm$15.07 & 27.20$\pm$11.61 & 71.30$\pm$0.46 & 26.77$\pm$5.68 & \textbf{25.82$\pm$5.73} & \textbf{79.78$\pm$0.39} & \textbf{70.86$\pm$5.55} & \textbf{63.91$\pm$8.26} & 69.70$\pm$0.19 & 70.50$\pm$0.31 & 68.14$\pm$1.10\\
 & CombNorm & 80.00$\pm$0.86 & 55.64$\pm$13.23 & 21.44$\pm$7.52 & 68.59$\pm$1.03 & 18.90$\pm$2.17 & 18.53$\pm$1.68 & 78.11$\pm$0.76 & 40.93$\pm$0.29 & 40.90$\pm$0.28  & \textbf{70.71$\pm$0.21} & \textbf{71.77$\pm$0.31} & \textbf{69.91$\pm$0.64}\\
 & NoNorm & 82.43$\pm$0.33 & 21.78$\pm$3.32 & 21.21$\pm$3.48 & \textbf{71.40$\pm$0.48} & 19.78$\pm$1.95 & 19.85$\pm$2.00 & 79.75$\pm$0.33 & 39.18$\pm$1.38 & 39.00$\pm$1.59 & 69.45$\pm$0.27 & 67.99$\pm$0.31 & 46.38$\pm$3.87\\
\midrule
\multirow{5}{*}{SGC}
 & BatchNorm & 79.32$\pm$0.76 & 15.86$\pm$8.86 & 14.40$\pm$7.02 & 61.60$\pm$1.31 & 17.34$\pm$4.05 & 17.82$\pm$3.86 & 76.34$\pm$0.75 & 54.22$\pm$15.63 & 29.49$\pm$11.83 & \textbf{68.58$\pm$0.15} & 65.54$\pm$0.38 & 62.33$\pm$0.61\\
 & PairNorm & 80.78$\pm$0.23 & 71.26$\pm$2.12 & 51.03$\pm$3.25 & 69.76$\pm$0.33 & 60.14$\pm$2.40 & 50.94$\pm$4.76 & 75.81$\pm$0.34 & 68.89$\pm$0.49 & 62.14$\pm$4.33 & 60.72$\pm$0.03 & 39.69$\pm$2.36 & 26.67$\pm$10.82\\
 & NodeNorm & 78.09$\pm$0.94 & \textbf{78.77$\pm$0.96} & 73.93$\pm$2.40 & 63.42$\pm$1.45 & 61.81$\pm$1.94 & 60.22$\pm$2.22 & 71.64$\pm$1.42 & 71.50$\pm$1.72 & 73.30$\pm$1.52 & 63.21$\pm$0.06 & 26.81$\pm$1.05 & 16.18$\pm$8.56\\
 & MeanNorm & 80.22$\pm$0.43 & 48.29$\pm$8.55 & 30.07$\pm$8.07 & 70.78$\pm$0.75 & 38.27$\pm$5.79 & 28.27$\pm$5.77 & 75.07$\pm$0.44 & 47.29$\pm$6.02 & 41.32$\pm$1.26 & 54.86$\pm$0.03 & 21.74$\pm$9.95 & 18.97$\pm$9.52\\
 & GroupNorm & \textbf{82.81$\pm$0.27} & 75.81$\pm$1.02 & \textbf{ 74.94$\pm$1.87 } & 72.32$\pm$0.43 & 67.54$\pm$0.78 & 61.75$\pm$0.76 & \textbf{78.87$\pm$0.50} & \textbf{76.43$\pm$1.03} & \textbf{74.62$\pm$1.29} & 66.12$\pm$0.04 & \textbf{67.29$\pm$0.53} & \textbf{66.11$\pm$0.46}\\
 & CombNorm & 77.65$\pm$1.68 & 75.16$\pm$2.27 & 74.45$\pm$1.86 & 63.66$\pm$1.41 & 59.97$\pm$3.90 & 54.52$\pm$4.79 & 71.67$\pm$2.30 & 71.50$\pm$2.27 & 72.23$\pm$2.46 & 65.73$\pm$0.07 & 54.37$\pm$0.53 & 47.52$\pm$1.40\\\
 & NoNorm & 79.38$\pm$0.39 & 75.93$\pm$1.12 & 68.75$\pm$2.59 & \textbf{72.36$\pm$0.39} & \textbf{71.06$\pm$1.32} & \textbf{62.64$\pm$3.68} & 78.01$\pm$0.32 & 69.06$\pm$0.57 & 66.55$\pm$0.56 & 61.96$\pm$0.07 & 41.43$\pm$0.25 & 34.24$\pm$0.07\\
\bottomrule
\end{tabular}}
\vspace{-2mm}
\end{table}

\textbf{More results of random droppings.} Table~\ref{tab:random_dropout_full} shows the performance of diverse random dropping tricks, where all dropout rates are tuned for best test accuarcies. Table~\ref{tab:random_dropout_0.2} and~\ref{tab:random_dropout_0.5} report the achieved performance with fixed dropout rates $0.2$ and $0.5$ respectively, for better comparisons. For all experiments, the standard deviations of $100$ independent repetitions are provided.

\vspace{-4mm}
\begin{table}[!ht]
\caption{Test accuracy (\%) with random dropping.}
\vspace{-4mm}
\label{tab:random_dropout_full}
\centering
\resizebox{0.48\textwidth}{!}{
\begin{tabular}{@{}llcccccccccccc@{}}
\toprule
\multirow{2}{*}{Dataset} & \multirow{2}{*}{Settings} & \multicolumn{3}{c}{Cora} & \multicolumn{3}{c}{Citeseer} & \multicolumn{3}{c}{PubMed} & \multicolumn{3}{c}{OGBN-ArXiv} \\
\cmidrule(r){3-5} \cmidrule(r){6-8} \cmidrule(r){9-11} \cmidrule(r){12-14}
& & 2 & 16 & 32 & 2 & 16 & 32 & 2 & 16 & 32 & 2 & 16 & 32 \\ 
\midrule
\multirow{8}{*}{GCN} & No Dropout & 80.68$\pm$0.13 & \textbf{28.56$\pm$2.77} & \textbf{29.36$\pm$2.76} & 71.36$\pm$0.18 & 23.19$\pm$1.47 & \textbf{23.03$\pm$1.13} & 79.56$\pm$0.17 & 39.85$\pm$1.47 & 40.00$\pm$1.59 & \textbf{69.53$\pm$0.19} & 66.14$\pm$0.73 & 41.96$\pm$9.01 \\
& Dropout & \textbf{82.39$\pm$0.36} & 21.60$\pm$3.54 & 21.17$\pm$3.29 & \textbf{71.43$\pm$0.46} & 19.37$\pm$1.88 & 20.15$\pm$1.85 & \textbf{79.79$\pm$0.37} & 39.09$\pm$1.23 & 39.17$\pm$1.43 & 69.40$\pm$0.12 & 67.79$\pm$0.54 & 45.41$\pm$3.63 \\
& DropNode & 77.10$\pm$1.04 & 27.61$\pm$4.34 & 27.65$\pm$5.07 & 69.38$\pm$0.89 & 21.83$\pm$3.07 & 22.18$\pm$3.06 & 77.39$\pm$0.98 & 40.31$\pm$1.61 & 40.38$\pm$1.20 & 66.67$\pm$0.16 & 67.17$\pm$0.44 & 43.81$\pm$9.62 \\
& DropEdge & 79.16$\pm$0.73 & 28.00$\pm$3.36 & 27.87$\pm$3.04 & 70.26$\pm$0.70 & 22.92$\pm$1.95 & 22.92$\pm$2.12 & 78.58$\pm$0.58 & 40.61$\pm$1.20 & 40.50$\pm$1.48 & 68.67$\pm$0.17 & 66.50$\pm$0.39 & \textbf{51.70$\pm$5.08} \\
& LADIES & 77.12$\pm$0.82 & 28.07$\pm$5.07 & 27.54$\pm$4.19 & 68.87$\pm$0.84 & 22.52$\pm$2.42 & 22.60$\pm$2.23 & 78.31$\pm$0.82 & 40.07$\pm$1.49 & 40.11$\pm$1.36 & 66.43$\pm$0.21 & 62.05$\pm$0.80 & 40.41$\pm$6.22 \\
& DropNode+Dropout & 81.02$\pm$0.75 & 22.24$\pm$8.30 & 18.81$\pm$4.30 & 70.59$\pm$0.80 & 24.49$\pm$6.50 & 18.23$\pm$0.55 & 78.85$\pm$0.64 & 40.44$\pm$1.22 & 40.37$\pm$1.36 & 68.66$\pm$0.14 & 68.27$\pm$0.33 & 44.18$\pm$4.90 \\
& DropEdge+Dropout & 79.71$\pm$0.92 & 20.45$\pm$7.97 & 21.10$\pm$8.17 & 69.64$\pm$0.89 & 19.77$\pm$4.47 & 18.49$\pm$1.71 & 77.77$\pm$0.99 & 40.71$\pm$0.84 & 40.51$\pm$0.96 & 66.55$\pm$0.14 & \textbf{68.81$\pm$0.21} & 49.82$\pm$4.16 \\
& LADIES+Dropout & 78.88$\pm$0.79 & 19.49$\pm$8.31 & 16.92$\pm$6.60 & 69.02$\pm$0.88 & \textbf{27.17$\pm$6.74} & 18.54$\pm$2.38 & 78.53$\pm$0.77 & \textbf{41.43$\pm$2.59} & \textbf{40.70$\pm$1.00} & 66.35$\pm$0.17 & 65.13$\pm$0.33 & 39.99$\pm$4.74 \\
\midrule
\multirow{8}{*}{SGC} & No Dropout & 77.55$\pm$0.08 & 73.99$\pm$0.03 & 66.80$\pm$0.01 & 71.80$\pm$0.00 & \textbf{72.69$\pm$0.05} & 70.50$\pm$0.02 & 77.59$\pm$0.03 & 69.74$\pm$0.07 & 67.81$\pm$0.03 & \textbf{62.34$\pm$0.07} & \textbf{42.54$\pm$0.25} & \textbf{34.76$\pm$0.06} \\
& Dropout & 79.37$\pm$0.36 & 75.91$\pm$1.07 & 68.40$\pm$2.57 & \textbf{72.35$\pm$0.39} & 71.21$\pm$0.93 & 62.35$\pm$3.15 & 78.04$\pm$0.33 & 69.12$\pm$0.54 & 66.53$\pm$0.68 & 61.96$\pm$0.05 & 41.47$\pm$0.23 & 34.22$\pm$0.06 \\
& DropNode & 78.57$\pm$0.27 & 76.99$\pm$0.31 & \textbf{72.93$\pm$0.39} & 71.87$\pm$0.27 & 72.50$\pm$0.20 & \textbf{70.60$\pm$0.11} & 77.63$\pm$0.32 & 72.51$\pm$0.29 & 68.16$\pm$0.33 & 61.21$\pm$0.08 & 40.52$\pm$0.11 & 34.64$\pm$0.05 \\
& DropEdge & 78.68$\pm$0.24 & 70.65$\pm$0.80 & 44.00$\pm$0.90 & 71.94$\pm$0.32 & 69.43$\pm$0.57 & 45.13$\pm$0.93 & \textbf{78.26$\pm$0.32} & 68.39$\pm$0.26 & 52.08$\pm$0.79 & 62.06$\pm$0.05 & 41.03$\pm$0.23 & 33.61$\pm$0.06 \\
& LADIES & 78.50$\pm$0.27 & \textbf{78.35$\pm$0.34} & 72.71$\pm$0.84 & 71.88$\pm$0.28 & 71.69$\pm$0.43 & 69.80$\pm$0.36 & 77.65$\pm$0.23 & \textbf{74.86$\pm$0.88} & \textbf{72.27$\pm$0.55} & 61.49$\pm$0.07 & 38.96$\pm$0.08 & 33.17$\pm$0.04 \\
& DropNode+Dropout & \textbf{80.60$\pm$0.41} & 74.83$\pm$1.64 & 55.04$\pm$4.28 & 72.33$\pm$0.43 & 70.30$\pm$1.56 & 65.85$\pm$1.86 & 78.10$\pm$0.41 & 67.98$\pm$0.67 & 52.01$\pm$1.36 & 61.54$\pm$0.05 & 39.48$\pm$0.13 & 32.63$\pm$0.07 \\
& DropEdge+Dropout & 80.27$\pm$0.50 & 76.19$\pm$1.28 & 66.08$\pm$3.55 & 72.09$\pm$0.45 & 66.48$\pm$3.38 & 35.55$\pm$2.79 & 77.63$\pm$0.37 & 69.65$\pm$0.67 & 67.55$\pm$0.79 & 60.21$\pm$0.07 & 39.12$\pm$0.10 & 33.81$\pm$0.06 \\
& LADIES+Dropout & 79.81$\pm$0.56 & 74.72$\pm$1.34 & 66.62$\pm$2.49 & 71.85$\pm$0.39 & 69.24$\pm$1.83 & 50.81$\pm$4.09 & 77.46$\pm$0.33 & 70.54$\pm$0.36 & 67.94$\pm$0.63 & 60.27$\pm$0.07 & 31.41$\pm$0.03 & 24.86$\pm$1.15 \\
\bottomrule
\end{tabular}}
\vspace{-4mm}
\end{table}

\vspace{-4mm}
\begin{table}[!ht]
\caption{Test accuracy (\%) with random dropping $p=0.2$.}
\vspace{-4mm}
\label{tab:random_dropout_0.2}
\centering
\resizebox{0.48\textwidth}{!}{
\begin{tabular}{@{}llcccccccccccc@{}}
\toprule
\multirow{2}{*}{Dataset} & \multirow{2}{*}{Settings} & \multicolumn{3}{c}{Cora} & \multicolumn{3}{c}{Citeseer} & \multicolumn{3}{c}{PubMed} & \multicolumn{3}{c}{OGBN-ArXiv} \\
\cmidrule(r){3-5} \cmidrule(r){6-8} \cmidrule(r){9-11} \cmidrule(r){12-14}
& & 2 & 16 & 32 & 2 & 16 & 32 & 2 & 16 & 32 & 2 & 16 & 32 \\ 
\midrule
\multirow{10}{*}{GCN} & DropNode ($p=0.2$) & 77.10$\pm$1.04 & 27.61$\pm$4.34 & 27.65$\pm$5.07 & 69.38$\pm$0.89 & 21.83$\pm$3.07 & 22.18$\pm$3.06 & 77.39$\pm$0.98 & 40.18$\pm$1.18 & 39.88$\pm$2.54 & 66.67$\pm$0.16 & 67.17$\pm$0.47 & 43.81$\pm$9.62 \\
& DropEdge ($p=0.2$) & 79.16$\pm$0.73 & 28.00$\pm$3.36 & 27.87$\pm$3.04 & 70.26$\pm$0.70 & 22.92$\pm$1.95 & 22.92$\pm$2.12 & 78.58$\pm$0.58 & 40.32$\pm$1.63 & 40.21$\pm$1.65 & 68.67$\pm$0.17 & 66.38$\pm$0.60 & 45.74$\pm$5.65 \\
& LADIES ($p=0.2$) & 77.12$\pm$0.82 & 28.07$\pm$5.07 & 27.54$\pm$4.19 & 68.87$\pm$0.84 & 22.52$\pm$2.42 & 22.60$\pm$2.23 & 78.31$\pm$0.82 & 39.98$\pm$1.76 & 39.82$\pm$1.63 & 66.43$\pm$0.21 & 62.05$\pm$0.80 & 40.41$\pm$6.22 \\
& DropNode+Dropout ($p=0.2$) & 81.02$\pm$0.75 & 18.55$\pm$4.68 & 18.81$\pm$4.30 & 70.59$\pm$0.80 & 20.03$\pm$4.68 & 18.16$\pm$0.79 & 78.85$\pm$0.64 & 40.22$\pm$1.60 & 39.92$\pm$1.73 & 68.66$\pm$0.14 & 68.19$\pm$0.25 & 44.18$\pm$4.90 \\
& DropEdge+Dropout ($p=0.2$) & 79.71$\pm$0.92 & 17.83$\pm$4.80 & 19.32$\pm$5.60 & 69.64$\pm$0.89 & 18.48$\pm$1.87 & 18.49$\pm$1.71 & 77.77$\pm$0.99 & 39.92$\pm$1.29 & 39.75$\pm$1.28 & 66.55$\pm$0.14 & 68.79$\pm$0.34 & 49.82$\pm$4.16 \\
& LADIES+Dropout ($p=0.2$) & 78.88$\pm$0.79 & 19.49$\pm$8.31 & 16.92$\pm$6.60 & 69.02$\pm$0.88 & 27.17$\pm$6.74 & 18.54$\pm$2.38 & 78.53$\pm$0.77 & 40.01$\pm$2.52 & 39.92$\pm$1.36 & 66.35$\pm$0.17 & 65.13$\pm$0.33 & 39.99$\pm$4.74 \\
\midrule
\multirow{10}{*}{SGC} & DropNode ($p=0.2$) & 77.89$\pm$0.23 & 75.22$\pm$0.22 & 69.51$\pm$0.40 & 71.87$\pm$0.27 & 72.50$\pm$0.20 & 70.60$\pm$0.11 & 77.63$\pm$0.32 & 70.28$\pm$0.21 & 68.16$\pm$0.33 & 61.21$\pm$0.08 & 40.52$\pm$0.11 & 34.64$\pm$0.05 \\
& DropEdge ($p=0.2$) & 78.16$\pm$0.24 & 70.65$\pm$0.80 & 44.00$\pm$0.90 & 71.94$\pm$0.32 & 69.43$\pm$0.57 & 45.13$\pm$0.93 & 78.26$\pm$0.32 & 68.39$\pm$0.26 & 52.08$\pm$0.79 & 62.06$\pm$0.05 & 41.03$\pm$0.23 & 33.61$\pm$0.06 \\
& LADIES ($p=0.2$) & 77.85$\pm$0.28 & 76.93$\pm$0.39 & 72.14$\pm$0.38 & 71.88$\pm$0.28 & 71.69$\pm$0.43 & 69.80$\pm$0.36 & 77.65$\pm$0.23 & 72.46$\pm$0.40 & 69.93$\pm$0.46 & 61.49$\pm$0.07 & 38.96$\pm$0.08 & 33.17$\pm$0.04 \\
& DropNode+Dropout ($p=0.2$) & 79.95$\pm$0.49 & 74.83$\pm$1.64 & 55.04$\pm$4.28 & 72.33$\pm$0.43 & 70.30$\pm$1.56 & 65.85$\pm$1.86 & 78.10$\pm$0.41 & 67.98$\pm$0.67 & 52.01$\pm$1.36 & 61.54$\pm$0.05 & 39.48$\pm$0.13 & 32.63$\pm$0.07 \\
& DropEdge+Dropout ($p=0.2$) & 80.11$\pm$0.38 & 76.19$\pm$1.28 & 66.08$\pm$3.55 & 72.09$\pm$0.45 & 66.48$\pm$3.38 & 35.55$\pm$2.79 & 77.63$\pm$0.37 & 69.65$\pm$0.67 & 67.55$\pm$0.79 & 60.21$\pm$0.07 & 39.12$\pm$0.10 & 33.81$\pm$0.06 \\
& LADIES+Dropout ($p=0.2$) & 79.63$\pm$0.44 & 74.72$\pm$1.34 & 66.62$\pm$2.49 & 71.85$\pm$0.39 & 69.24$\pm$1.83 & 50.81$\pm$4.09 & 77.46$\pm$0.33 & 69.81$\pm$0.82 & 67.94$\pm$0.63 & 60.27$\pm$0.07 & 31.41$\pm$0.03 & 24.86$\pm$1.15 \\
\bottomrule
\end{tabular}}
\vspace{-2mm}
\end{table}

\textbf{More results of comparison with previous state-of-the-art frameworks.} We present a more complete comparison with other previous state-of-the-art frameworks of training deep GNNs in Table~\ref{tab:model_comparison_full}. Error bars are also recorded. 

\begin{table}[!ht]
\caption{Test accuracy (\%) with random dropping $p=0.5$.}
\vspace{-4mm}
\label{tab:random_dropout_0.5}
\centering
\resizebox{0.48\textwidth}{!}{
\begin{tabular}{@{}llcccccccccccc@{}}
\toprule
\multirow{2}{*}{Dataset} & \multirow{2}{*}{Settings} & \multicolumn{3}{c}{Cora} & \multicolumn{3}{c}{Citeseer} & \multicolumn{3}{c}{PubMed} & \multicolumn{3}{c}{OGBN-ArXiv} \\
\cmidrule(r){3-5} \cmidrule(r){6-8} \cmidrule(r){9-11} \cmidrule(r){12-14}
& & 2 & 16 & 32 & 2 & 16 & 32 & 2 & 16 & 32 & 2 & 16 & 32 \\ 
\midrule
\multirow{10}{*}{GCN} & DropNode ($p=0.5$) & 70.28$\pm$1.36 & 21.09$\pm$5.86 & 20.62$\pm$5.92 & 65.47$\pm$1.31 & 18.68$\pm$3.05 & 19.38$\pm$2.84 & 73.82$\pm$1.38 & 40.31$\pm$1.61 & 40.38$\pm$1.20 & 60.64$\pm$0.16 & 67.17$\pm$0.44 & 35.65$\pm$14.39 \\
& DropEdge ($p=0.5$) & 75.94$\pm$1.19 & 25.38$\pm$4.20 & 25.27$\pm$4.37 & 68.06$\pm$1.08 & 21.06$\pm$2.68 & 21.34$\pm$2.53 & 76.33$\pm$1.03 & 40.61$\pm$1.20 & 40.50$\pm$1.48 & 66.86$\pm$0.23 & 66.50$\pm$0.39 & 51.70$\pm$5.08 \\
& LADIES ($p=0.5$) & 70.59$\pm$1.15 & 24.57$\pm$7.84 & 25.77$\pm$7.43 & 65.19$\pm$1.39 & 20.64$\pm$2.98 & 19.98$\pm$2.67 & 75.32$\pm$1.39 & 40.07$\pm$1.49 & 40.11$\pm$1.36 & 60.46$\pm$0.19 & 55.86$\pm$0.48 & 34.11$\pm$6.28 \\
& DropNode+Dropout ($p=0.5$) & 78.30$\pm$1.00 & 22.24$\pm$8.30 & 18.74$\pm$6.68 & 68.20$\pm$1.05 & 24.49$\pm$6.50 & 18.23$\pm$0.55 & 76.75$\pm$1.02 & 40.44$\pm$1.22 & 40.37$\pm$1.36 & 66.82$\pm$0.20 & 68.27$\pm$0.33 & 42.48$\pm$4.50 \\
& DropEdge+Dropout ($p=0.5$) & 72.31$\pm$1.21 & 20.45$\pm$7.97 & 21.10$\pm$8.17 & 64.69$\pm$1.35 & 19.77$\pm$4.47 & 18.22$\pm$0.90 & 75.21$\pm$1.04 & 40.71$\pm$0.84 & 40.51$\pm$0.96 & 60.17$\pm$0.14 & 68.81$\pm$0.21 & 47.03$\pm$8.06 \\
& LADIES+Dropout ($p=0.5$) & 71.65$\pm$1.13 & 17.82$\pm$7.84 & 16.79$\pm$7.79 & 64.28$\pm$1.36 & 23.63$\pm$5.84 & 18.05$\pm$1.48 & 75.89$\pm$1.34 & 41.43$\pm$2.59 & 40.70$\pm$1.00 & 59.96$\pm$0.22 & 59.21$\pm$0.40 & 30.21$\pm$1.94 \\
\midrule
\multirow{10}{*}{SGC} & DropNode ($p=0.5$) & 78.57$\pm$0.27 & 76.99$\pm$0.31 & 72.93$\pm$0.39 & 70.84$\pm$0.32 & 71.59$\pm$0.20 & 70.54$\pm$0.28 & 77.38$\pm$0.22 & 72.51$\pm$0.29 & 68.08$\pm$0.39 & 58.29$\pm$0.08 & 38.80$\pm$0.06 & 33.92$\pm$0.03 \\
& DropEdge ($p=0.5$) & 78.68$\pm$0.24 & 63.92$\pm$1.59 & 29.50$\pm$1.66 & 71.83$\pm$0.26 & 57.23$\pm$1.45 & 19.97$\pm$0.48 & 77.95$\pm$0.31 & 65.97$\pm$1.24 & 48.38$\pm$1.69 & 60.93$\pm$0.08 & 37.01$\pm$0.18 & 28.30$\pm$0.70 \\
& LADIES ($p=0.5$) & 78.50$\pm$0.27 & 78.35$\pm$0.34 & 72.71$\pm$0.84 & 69.98$\pm$0.29 & 70.66$\pm$0.44 & 69.24$\pm$0.39 & 77.27$\pm$0.34 & 74.86$\pm$0.88 & 72.27$\pm$0.55 & 58.89$\pm$0.09 & 35.78$\pm$0.05 & 31.35$\pm$0.02 \\
& DropNode+Dropout ($p=0.5$) & 80.60$\pm$0.41 & 68.37$\pm$2.54 & 8.14$\pm$0.72 & 72.08$\pm$0.61 & 63.33$\pm$3.27 & 48.22$\pm$2.64 & 77.94$\pm$0.35 & 64.55$\pm$3.05 & 44.72$\pm$0.61 & 59.79$\pm$0.09 & 34.28$\pm$0.06 & 27.42$\pm$0.56 \\
& DropEdge+Dropout ($p=0.5$) & 80.27$\pm$0.50 & 59.52$\pm$7.94 & 22.44$\pm$8.83 & 70.86$\pm$0.46 & 27.48$\pm$6.40 & 18.12$\pm$0.30 & 77.37$\pm$0.47 & 68.67$\pm$1.43 & 64.27$\pm$1.78 & 55.73$\pm$0.07 & 36.29$\pm$0.05 & 31.66$\pm$0.03 \\
& LADIES+Dropout ($p=0.5$) & 79.81$\pm$0.56 & 67.00$\pm$3.14 & 42.08$\pm$6.53 & 70.25$\pm$0.49 & 46.27$\pm$4.73 & 34.90$\pm$0.58 & 76.88$\pm$0.45 & 70.54$\pm$0.36 & 67.86$\pm$0.99 & 56.02$\pm$0.08 & 27.71$\pm$0.01 & 15.65$\pm$7.59 \\
\bottomrule
\end{tabular}}
\vspace{-4mm}
\end{table}

\begin{table}[!ht]
\caption{{\small Acc. (\%) ($100$ runs) comparison with previous SOTAs.}}
\vspace{-4mm}
\label{tab:model_comparison_full}
\centering
\resizebox{0.48\textwidth}{!}{
\begin{tabular}{@{}lcccccccccccc@{}}
\toprule
\multirow{2}{*}{Model}  & \multicolumn{3}{c}{Cora (\textbf{Ours: 85.48$\pm$0.39})} & \multicolumn{3}{c}{Citeseer (\textbf{Ours: 73.35$\pm$0.79})} & \multicolumn{3}{c}{PubMed (\textbf{Ours: 80.76$\pm$0.26})} & \multicolumn{3}{c}{OGBN-ArXiv (\textbf{Ours: 72.70$\pm$0.15})}\\
\cmidrule(r){2-4} \cmidrule(r){5-7} \cmidrule(r){8-10} \cmidrule(r){11-13}
& 2 & 16 & 32 & 2 & 16 & 32 & 2 & 16 & 32 & 2 & 16 & 32\\ 
\midrule
SGC~\cite{wu2019simplifying} & 79.31$\pm$0.37 & 75.98$\pm$1.06 & 68.45$\pm$3.10 & 72.31$\pm$0.38 & 71.03$\pm$1.18 & 61.92$\pm$3.48 & 78.06$\pm$0.31 & 69.18$\pm$0.58 & 66.61$\pm$0.56 & 61.98$\pm$0.08 & 41.58$\pm$0.27 & 34.22$\pm$0.04\\
DAGNN~\cite{liu2020towards} & 80.30$\pm$0.78 & 84.14$\pm$0.59 & 83.39$\pm$0.59 & 18.22$\pm$3.48 & 73.05$\pm$0.62 & 72.59$\pm$0.54 & 77.74$\pm$0.57 & 80.32$\pm$0.38 & 80.58$\pm$0.51 & 67.65$\pm$0.52 & 71.82$\pm$0.28 & 71.46$\pm$0.27  \\
GCNII~\cite{chen2020simple} & 82.19$\pm$0.77 & 84.69$\pm$0.51 & 85.29$\pm$0.47 & 67.81$\pm$0.89 & 72.97$\pm$0.71 & 73.24$\pm$0.78 & 78.05$\pm$1.53 & 80.03$\pm$0.50 & 79.91$\pm$0.27 & 71.24$\pm$0.17 & 72.61$\pm$0.29 & 72.60$\pm$0.25 \\
JKNet~\cite{xu2018representation} & 79.06$\pm$0.11 & 72.97$\pm$3.94 & 73.23$\pm$3.59 & 66.98$\pm$1.82 & 54.33$\pm$7.74 & 50.68$\pm$8.73 & 77.24$\pm$0.92 & 64.37$\pm$8.80 & 63.77$\pm$9.21 & 63.73$\pm$0.38 & 66.41$\pm$0.56 & 66.31$\pm$0.63  \\
APPNP~\cite{klicpera2018predict} & 82.06$\pm$0.46 & 83.64$\pm$0.48 & 83.68$\pm$0.48 & 71.67$\pm$0.78 & 72.13$\pm$0.53 & 72.13$\pm$0.59 & 79.46$\pm$0.47 & 80.30$\pm$0.30 & 80.24$\pm$0.33 & 65.31$\pm$0.23 & 66.95$\pm$0.24 & 66.94$\pm$0.26  \\
GPRGNN~\cite{chien2021adaptive} & 82.53$\pm$0.49 & 83.69$\pm$0.55 & 83.13$\pm$0.60 & 70.49$\pm$0.95 & 71.39$\pm$0.73 & 71.01$\pm$0.79 & 78.73$\pm$0.63 & 78.78$\pm$1.02 & 78.46$\pm$1.03 & 69.31$\pm$0.09 & 70.30$\pm$0.15 & 70.18$\pm$0.16 \\
\bottomrule
\end{tabular}}
\vspace{-4mm}
\end{table}

\begin{table}[!ht]
\caption{\small Cora with $2/4/8/16/32/64/128$ GNNs from $100$ runs.}
\vspace{-4mm}
\label{tab:complete_cora}
\centering
\resizebox{0.48\textwidth}{!}{
\begin{tabular}{lcccccccc}
\toprule
\multirow{2}{*}{Model}  & \multicolumn{8}{c}{Cora (\textbf{Ours: 85.48})} \\
\cmidrule{2-9}
& 2 & 4 & 8 & 16 & 24 & 32 & 64 & 128 \\ 
\midrule
SGC & 79.31$\pm$0.37 & 77.34$\pm$0.26 & 74.94$\pm$0.31 & 75.98$\pm$1.06 & 68.34$\pm$0.88 &  68.45$\pm$3.10 & 25.27$\pm$1.79 & 20.15$\pm$2.21\\
DAGNN & 80.30$\pm$0.78 & 83.50$\pm$0.57 & 84.36$\pm$0.46 & 84.14$\pm$0.59 & 83.56$\pm$0.54 & 83.39$\pm$0.59 & 82.34$\pm$0.73 & 78.78$\pm$2.22\\
GCNII & 82.19$\pm$0.77 & 82.84$\pm$0.61 & 84.07$\pm$0.64 & 84.69$\pm$0.51 &  85.02$\pm$0.49 & 85.29$\pm$0.47 & 85.23$\pm$0.55 & 85.28$\pm$0.49\\
JKNet & 79.06$\pm$0.11 & 75.71$\pm$2.85 & 74.62$\pm$2.83 & 72.97$\pm$3.94 & 73.57$\pm$3.11 & 73.23$\pm$3.59 & 71.66$\pm$5.36 & 70.93$\pm$6.91\\
APPNP & 82.06$\pm$0.46 & 83.17$\pm$0.42 & 83.51$\pm$0.49 & 83.64$\pm$0.48 & 83.69$\pm$0.50 & 83.68$\pm$0.48 & 83.67$\pm$0.51 & 83.64$\pm$0.48\\
GPRGNN& 82.53$\pm$0.49 & 83.45$\pm$0.50 & 83.79$\pm$0.56 & 83.69$\pm$0.55 & 83.28$\pm$0.64 & 83.13$\pm$0.60 & 82.34$\pm$1.34 & 81.93$\pm$2.09 \\
Ours & 83.59$\pm$0.37 & 84.42$\pm$0.37 & 85.11$\pm$0.41 & 85.38$\pm$0.53  & 85.40$\pm$0.47 & 85.39$\pm$0.43 & \textbf{85.48$\pm$0.42} & 85.46$\pm$0.40 \\
\bottomrule
\end{tabular}}
\vspace{-4mm}
\end{table}

\begin{table}[!ht]
\caption{\small Citeseer with $2/4/8/16/32/64/128$ GNNs from $100$ runs.}
\vspace{-4mm}
\label{tab:complete_citeseer}
\centering
\resizebox{0.48\textwidth}{!}{
\begin{tabular}{lcccccccc}
\toprule
\multirow{2}{*}{Model}  & \multicolumn{8}{c}{Citeseer (\textbf{Ours: 73.35})} \\
\cmidrule{2-9}
& 2 & 4 & 8 & 16 & 24 & 32 & 64 & 128 \\ 
\midrule
SGC & 72.31$\pm$0.38 & 68.67$\pm$0.32 & 68.73$\pm$0.41 & 71.03$\pm$1.18 & 66.99$\pm$0.40 & 61.92$\pm$3.48 & 63.07$\pm$0.42 & 54.86$\pm$0.44\\
DAGNN & 18.22$\pm$3.48 & 19.73$\pm$5.25  & 73.19$\pm$0.64 & 73.05$\pm$0.62 & 72.75$\pm$0.62 & 72.59$\pm$0.54 & 71.68$\pm$0.65 & 70.91$\pm$0.76\\
GCNII & 67.81$\pm$0.89 & 68.10$\pm$0.84 & 70.80$\pm$0.69  & 72.97$\pm$0.71 & 73.22$\pm$0.85 &  73.24$\pm$0.78  & 73.16$\pm$0.85 & 72.84$\pm$0.93 \\
JKNet & 66.98$\pm$1.82 & 60.21$\pm$5.89 & 56.33$\pm$7.91 &  54.33$\pm$7.74 & 51.57$\pm$8.08 & 50.68$\pm$8.73 & 48.99$\pm$8.15 & 47.53$\pm$8.81\\
APPNP & 71.67$\pm$0.78 & 71.95$\pm$0.64 & 72.09$\pm$0.54 & 72.13$\pm$0.53 & 72.12$\pm$0.63 & 72.13$\pm$0.59 & 72.16$\pm$0.55 & 72.11$\pm$0.53\\
GPRGNN& 70.49$\pm$0.95 & 71.05$\pm$0.76 & 71.67$\pm$0.78 & 71.39$\pm$0.73 & 71.50$\pm$0.89 &  71.01$\pm$0.79 & 70.71$\pm$0.94 & 70.05$\pm$1.23\\
Ours & 68.00$\pm$0.81 & 68.45$\pm$0.73 & 72.22$\pm$0.65 & 72.97$\pm$0.62 & 73.16$\pm$0.61 & \textbf{73.35$\pm$0.83} & 73.17$\pm$0.89 & 73.00$\pm$0.94\\
\bottomrule
\end{tabular}}
\vspace{-4mm}
\end{table}

\begin{table}[!ht]
\caption{\small PubMed with $2/4/8/16/32/64/128$ GNNs $100$ runs.}
\vspace{-4mm}
\label{tab:complete_pubmed}
\centering
\resizebox{0.48\textwidth}{!}{
\begin{tabular}{lcccccccc}
\toprule
\multirow{2}{*}{Model}  & \multicolumn{8}{c}{Pubmed (\textbf{Ours: 80.76})} \\
\cmidrule{2-9}
& 2 & 4 & 8 & 16 & 24 & 32 & 64 & 128 \\ 
\midrule
SGC & 78.06$\pm$0.31 & 73.46$\pm$0.18 & 71.40$\pm$0.16 & 69.18$\pm$0.58 & 70.04$\pm$0.07 & 66.61$\pm$0.56 & 41.04$\pm$1.99 & 19.90$\pm$6.48\\
DAGNN & 77.74$\pm$0.57 & 79.00$\pm$0.53 & 79.84$\pm$0.46 & 80.32$\pm$0.38 & 80.55$\pm$0.40 & 80.58$\pm$0.51 & 80.44$\pm$0.48 & 79.74$\pm$0.64\\
GCNII & 78.05$\pm$1.53 & 77.86$\pm$0.91 & 78.09$\pm$1.09 & 80.03$\pm$0.50 & 79.88$\pm$0.33 & 79.91$\pm$0.27 & 79.98$\pm$0.26 & 79.91$\pm$0.30\\
JKNet & 77.24$\pm$0.92 & 71.21$\pm$5.11 & 65.09$\pm$8.28 & 64.37$\pm$8.80 & 62.86$\pm$8.73 & 63.77$\pm$9.21 & 62.97$\pm$8.96 & 62.31$\pm$8.49\\
APPNP & 79.46$\pm$0.47 & 79.82$\pm$0.32 & 80.16$\pm$0.30 & 80.30$\pm$0.30 & 80.25$\pm$0.31 & 80.24$\pm$0.33 & 80.26$\pm$0.29 & 80.27$\pm$0.28\\
GPRGNN& 78.73$\pm$0.63 & 78.71$\pm$0.80 & 78.51$\pm$1.04 & 78.78$\pm$1.02 & 78.08$\pm$1.18 & 78.46$\pm$1.03 & 79.04$\pm$1.12 & 79.80$\pm$0.61\\
Ours & 77.60$\pm$0.06 & 78.6$\pm$0.05 & 80.18$\pm$0.12 & 80.12$\pm$0.09 & 80.20$\pm$0.11 & \textbf{80.76$\pm$0.26} & 80.10$\pm$0.23 & 79.50$\pm$0.11\\
\bottomrule
\end{tabular}}
\vspace{-2mm}
\end{table}

\textbf{More results of depth studies.} To comprehensively validate the effectiveness of our explored best combo of tricks, we compare with the baselines of deep GNNs by considering a wide series of model depths, including $2$, $4$, $8$, $16$, $24$, $32$, $64$, and $128$. The node classification results on Cora, Citeseer, and Pubmed are shown in Tables. We find: 

(i) On Cora, Citesser, and Pubmed, our best trick combos consistently achieve the superior performances with larger model depths, compared to existing state-of-the-art GNNs. 

(ii) The traditional deep GNNs (e.g., SGC, JKNet, APPNP, GPRGNN) accompanied with the sub-optimal tricks often have weak performance gains against their shallow variants. In some cases, e.g., JKNet on the three citation datasets, the model performances even decrease with model depths. 

(iii) Our trick combos enable deep GNNs to surpass shallow GNNs by a substantial performance margins on all the classical graph datasets. In general, along with the increasing of depth, the performance of our deep GNNs first increases, then reaches the ``sweet depth" with the highest accuracy, and in the end starts to degrade. We observe the ``sweet depths" on \{Cora, Citeseer, PubMed\} are \{64,32,32\} layers. 

(iv) We could draw conclusion that the increasing model depth does benefit to GNN model, only under the condition of optimal trick combo. For each graph analytics problem, there should be a specific ``sweet depth" to access the highest performance. 

\end{document}

%% file: math_commands.tex
\usepackage{amsmath,amsfonts,bm}

\def\eqref#1{equation~\ref{#1}}

\def\1{\bm{1}}

\def\mE{{\bm{E}}}

\def\mR{{\bm{R}}}

\DeclareMathAlphabet{\mathsfit}{\encodingdefault}{\sfdefault}{m}{sl}
\SetMathAlphabet{\mathsfit}{bold}{\encodingdefault}{\sfdefault}{bx}{n}

